\documentclass{article}
\PassOptionsToPackage{numbers, compress}{natbib}
 \usepackage[preprint]{neurips_2026}

\usepackage[utf8]{inputenc} % allow utf-8 input
\usepackage[T1]{fontenc}    % use 8-bit T1 fonts
\usepackage{hyperref}       % hyperlinks
\usepackage{url}            % simple URL typesetting
\usepackage{booktabs}       % professional-quality tables
\usepackage{amsfonts}       % blackboard math symbols
\usepackage{nicefrac}       % compact symbols for 1/2, etc.
\usepackage{microtype}      % microtypography
\usepackage{xcolor}         % colors

\usepackage{listings}
\usepackage[most]{tcolorbox}
\tcbuselibrary{listings,skins,breakable,raster}
\usepackage{amsmath,amssymb}
\usepackage{wrapfig}
\usepackage{enumitem}

\usepackage{graphicx}
\usepackage{multirow}
\usepackage{makecell}
\usepackage[table]{xcolor}
\usepackage{capt-of}

\title{Attend to Your Own Thoughts: Breaking the Barrier for Post-Training Quantization of Reasoning LLMs through the Lens of 1.58-Bit Quantization}
\author{%
  Shigeng Wang\thanks{\ Equal contribution. This work was done when Shigeng Wang was an intern at Intel Labs China.} , Chao Li$^*$, Yangyuxuan Kang, Jiawei Fan, Anbang Yao\thanks{\ Corresponding author who conceived the project and led the writing of the paper.}\\
  Intel Labs China\\
  \texttt{\{shigeng.wang,chao3.li,yangyuxuan.kang,jiawei.fan,anbang.yao\}@intel.com} \\
}
\begin{document}

\maketitle

\begin{abstract}

While ternary (i.e., 1.58-bit) quantization can substantially reduce memory footprint and accelerate inference for LLMs, its adoption in real-world applications remains limited. This is primarily due to its susceptibility to severe performance degradation. Existing state-of-the-art methods mainly rely on quantization-aware training to mitigate this issue and thus incur prohibitive costs, scaling poorly across complex reasoning tasks, diverse model architectures and large-scale models. To address these limitations, we propose~\textbf{ScaleQ-1.58}, a scalable ternary post-training quantization (PTQ) framework for reasoning LLMs. Its core insight stems from an empirical finding: although modern LLMs are typically trained to exhibit chain-of-thought reasoning capabilities, in the PTQ regime, even the latest CAT-Q method based on learning-based differentiable ternarization still leads to performance collapse on challenging mathematics and coding tasks when using conventional calibration schemes that ignore the model's reasoning process. Driven by this finding, we introduce a simple calibration approach, \textbf{Attend to Your Own Thoughts} (\textbf{AYOT}), where reasoning traces and final answers generated by the pre-trained high-precision target LLM on a proper set of calibration samples are used as the context input during the ternarization process, along with the corresponding questions. ScaleQ-1.58 is formed by simply integrating AYOT with CAT-Q, which demonstrates several scaling properties: (1) with only 4M calibration tokens, Qwen3-1.7B ternarized by ScaleQ-1.58 reaches over 90.52\% of the performance of the prior best BitNet b1.58 2B4T averaged over 4 mathematics and coding tasks, and our ternary Qwen3-4B shows an absolute gain of 8.97\%, while requiring 1,000,000$\times$ fewer calibration tokens for quantization; (2) ScaleQ-1.58 generalizes well to both dense and MoE architectures, with performance improving as model scale increases (up to 235B parameters); (3) ScaleQ-1.58 demonstrates strong generalization across tasks of varying difficulty levels, including mathematics, coding and scientific logic reasoning, as well as commonsense reasoning and basic language generation; (4) its performance continues to improve as the number of calibration tokens increases. Notably, AYOT also exhibits strong generalization ability across other quantization bit-widths. Code will be available at https://github.com/IntelChina-AI/BitTern.

% In this paper, we present ScaleQ-1.58, a scalable post-training ternarization framework for large language molarge language models with chain-of-thought reasoning abilities. Our empirical study shows that conventional calibration methods fail to address this underexplored problem, leading to huge performance drops. Motivated by this, we introduce Look at Prompt and own Thoughts (LPoT), a new calibration method, which can significantly alleviate this issue. Given a set of calibration samples, the core idea of LPoT is simply concatenating the reasoning thoughts generated by the target FP16 model to their corresponding prompts as the context for generating layer activations during the ternarization process. By integrating LPoT with the recently proposed soft ternarization strategy, we have ScaleQ-1.58. It achieves excellent performance on a wide range of LLMs compared to state-of-the-art 1.58-bit models while requiring substantial fewer computational resources. With only 4M calibration tokens ($10^6$ times fewer than BiTNet 2B4T): (1) Qwen3-1.7B ternarized by ScaleQ-1.58 reaches over 90.52\% performance of BitNet 2B4T, while our ternary Qwen3-4B attains 8.97\% absolute gain, in terms of accuracy averaged over four math (Math-500 \& GSM8K) and code (HumanEval+ \& MBPP+) tasks; (2) when applying ScaleQ-1.58 to Ring-flash-2.0 having up to 100B parameter, we get a high-performance ternary model within 70 hours on 8 Nvidia H100-80G GPUs. Intriguingly, ScaleQ-1.58 exhibits an increasing performance trend when scaling up calibration tokens. 

\end{abstract}

\section{Introduction}
  
In recent years, the reasoning capability of large language models (LLMs) on challenging tasks such as mathematics and coding has been greatly improved by the advent of chain-of-thought (CoT) prompting~\citep{wei2022CoT, openai-o1, deepseek-r1}. Moreover, CoT enhances the benefits of scaling up model size and exploiting %harnessing 
advanced model architectures, resulting in improved performance across a wide range of tasks~\citep{gpt4, qwen3, gpt-5, gemini-2.5, gpt-4o, s1, kimi-k1.5, qwen3.5, deepseekv4}. Consequently, the evolution of LLMs has accelerated rapidly. However, deploying LLMs in real-world applications, particularly on resource-constrained platforms, remains constrained due to their high memory and computational costs. Post-training quantization (PTQ) offers a practically appealing solution to this problem, as it is %both 
low-cost and easy to implement. Despite significant advances in PTQ research for LLMs, most existing methods~\citep{q-bert, llm-int8, zeroquant, gptq, smoothquant, omniquant, awq, quip, quarot, spinquant, flatquant, ostquant, sliderquant} focus on conventional low-bit settings (e.g., 8-bit/4-bit quantization) and primarily evaluate their performance on basic language generation and commonsense reasoning tasks, largely overlooking %challenging 
complex reasoning tasks such as mathematics and coding that are more critical in practice~\citep{sliderquant}. A recent work~\citep{benchmark_qrllm} benchmarks the impact of quantization on reasoning LLMs, revealing that existing PTQ methods generally incur serious performance degradation when reducing model precision to 3-bit.

~\textit{In this paper, we study post-training quantization for reasoning LLMs through the lens of ternary (i.e., 1.58-bit) quantization from an unexplored perspective of scaling behavior}. The concept of ternary quantization was originally proposed in TWN~\citep{twn} to train compact convolutional neural networks (ConvNets) for computer vision tasks. It approximates the high-precision weights $\textbf{W}$ for a linear layer as $\textbf{W} \approx \alpha \textbf{T}$, where $\alpha$ is a scaling factor and $\textbf{T}$ %$\in \{-1, 0, 1\}$ 
denotes the corresponding ternary weights. The elements of $\textbf{T}$ are obtained by a hard ternarization function based on a weight threshold $\Delta$:
\begin{equation}
T_i =
  \begin{cases}
    1, & if \textbf{ } W_i > \Delta \\
    \;\;0, & if \textbf{ } |W_i| \leq \Delta\\ %\in [-\Delta,\Delta] \\
    -1, & if \textbf{ } W_i < -\Delta.
  \end{cases}
   \label{eq1:twn}
   %\vspace{- 0.in}
\end{equation}
By constraining model weights to $\{1,0,-1\}$, most of floating-point multiplication operations can be replaced by lower-cost integer additions and subtractions, thereby significantly reducing memory footprint and accelerating inference. %Recently, 
Because of this hardware-friendly advantage, increasing research efforts have been devoted to %leveraging ternary quantization to 
developing 1.58-bit LLMs for natural language processing tasks. TernaryBERT~\citep{ternarybert} makes the first attempt to train 1.58-bit small BERT models~\citep{bert} via fine-tuning for basic language understanding tasks, with the largest model comprising only 418M parameters. Due to its extremely low-bit nature, ternary quantization is highly susceptible to severe performance degradation. TernaryBERT uses knowledge distillation (KD)~\citep{tinybert, kd} to mitigate this issue.~\citep{bitnet-distillation, ternaryllm} extend this KD-based ternarization strategy with fine-tuning to relatively larger models up to 8B parameters, but still focus on basic tasks. %, but their focus remains limited to simpler rather than complex reasoning tasks. %In contrast to them, 
Unlike them, BitNet b1.58 series~\citep{bitnet-1.58, bitnetv2}, a highly influential, open-source family of ternary LLMs whose largest model has 7B parameters, are trained from scratch on a corpus of 100B tokens. Following BitNet b1.58 series, %~\citep{bitnet-1.58, bitnetv2}, several top-performing 
subsequent 1.58-bit LLM families, such as TriLM~\citep{spectra} (with its largest model of 3.9B parameters trained on 300B tokens) and Tequila~\citep{tequila} (with its largest model of 3B parameters trained on 10B tokens), are also built based on quantization-aware training (QAT). Although QAT %has achieved promising results on language models having up to 7B parameters, 
can alleviate performance degradation issue to a large extent by simulating weight ternarization during full-precision training, it requires substantial training resources, such as a vast amount of training tokens, large-scale GPU clusters, long training time and considerable power consumption. The demand for such training resources further grows with increasing model scale and architectural complexity. As a result, these existing QAT-based ternarization methods are limited to specific dense LLM architectures (e.g., Llama-like models~\citep{llama,llama2}) with relatively small model scales (typically below 10B parameters). Furthermore, in performance benchmarks, they neglect challenging reasoning tasks, such as mathematics and coding, akin to most conventional PTQ methods for LLMs. BitNet b1.58 2B4T~\citep{bitnet-2b4t}, a %pioneering 
1.58-bit reasoning LLM, comprising 2 billion parameters and trained from scratch on an extremely large %significantly expanded 
corpus of 4 trillion tokens, including text, mathematics, code and conversation data. It is the first 1.58-bit LLM capable of handling complex reasoning tasks. Despite a relatively lightweight 2B model scale, its training pipeline is both costly and complex, involving two-stage large-scale pre-training, supervised fine-tuning and direct preference optimization, along with meticulous hyperparameter tuning to prevent training instability. This restricts its scalability to much larger LLMs. 

Note that, according to Equation~\ref{eq1:twn}%defined above
, the strategies for estimating the scaling factor $\alpha$, the weight threshold $\Delta$, and gradients (since ternarization is non-differentiable) constitute three primitive components of ternary quantization from an optimization perspective. Following TWN~\citep{twn} tailored for ConvNets, all aforementioned ternarization methods for LLMs also adopt statistical strategies to determine %the scaling factor 
$\alpha$ and %the weight threshold 
$\Delta$ in a static manner, and employ straight-through estimator (STE)~\citep{bnn} to approximate %and backpropagate 
gradients during ternarization. %through the non-differentiable ternary operation. 
Instead, by extending learning-based quantization methodologies~\citep{xnor-net, ttq, lsq, dsq, ternaryllm, paretoq}%for ConvNets and LLMs
, %the recently proposed CAT-Q
the authors of~\citep{cat-q} recently present CAT-Q, the first differentiable ternarization method, which learns group-wise scaling factors and weight thresholds for\begin{figure}[htbp]
    \centering
    \vskip -0.05 in
    \centerline{\includegraphics[width=0.9\textwidth]{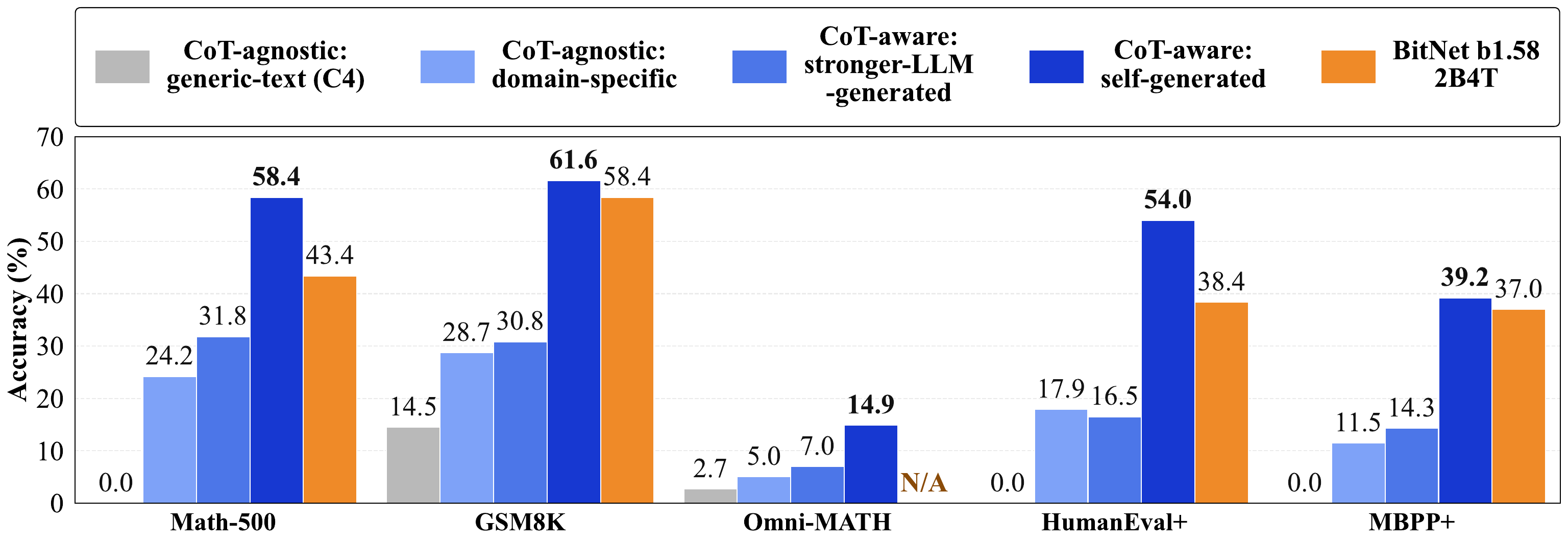}}
    \vskip -0.05 in
    \caption{Performance comparison of 1.58-bit Qwen3-4B models obtained via differentiable ternarization under different calibration schemes across five challenging mathematics and coding tasks. Using our calibration method, Attend to Your Own Thoughts (AYOT, i.e., CoT-aware: self-generated), with only 4M calibration tokens, the resulting 1.58-bit Qwen3-4B in the PTQ regime yields an absolute gain of 8.97\% (averaged over Math-500, GSM8K, HumanEval+ and MBPP+) compared to BitNet b1.58 2B4T (highlighted in orange), the prior best 1.58-bit reasoning LLM trained from scratch on %a massive amount of 
    4T tokens in the QAT regime, while achieving a reduction of 1,000,000$\times$ in training tokens. More accurate 1.58-bit LLMs spanning different architectures and model scales are reported in Table~\ref{tab:math_code_results}.}%Visualization of the calibration-data ablation in Table~\ref{tab:ablation_results}. Task-specific calibration substantially improves performance over generic calibration, and the self-generated calibration strategy yields the strongest overall results across mathematical reasoning and code generation benchmarks.}
    \label{fig01:empirical_study}
    \vskip -0.1 in
\end{figure}
pre-trained LLMs via a soft ternarization function. It achieves competitive performance at comparable model scales to QAT-based methods such as BitNet b1.58 series, TriLM, and Tequila, and demonstrates favorable scaling behavior as model scale increases, extending beyond 100B parameters. However, it still focuses on much simpler commonsense reasoning rather than challenging reasoning tasks. In short, the lack of a scalable ternary PTQ method for efficiently building diverse and high-performing 1.58-bit reasoning LLMs remains a key gap in the community.

%Motivated by the above analysis, this work %bridges this gap by 
\textit{To unlock the potential of %1.58-bit reasoning LLMs 
ternary quantization for reasoning LLMs and enable their broad applications and accessibility, this paper, based on CAT-Q, takes the first step toward investigating: in the PTQ regime}, (1) whether ternary quantization can scale across challenging reasoning tasks, diverse model architectures and larger model scales; (2) if not, what underlying factors underlie this bottleneck? % issue 
To explore these two questions, we adopt CAT-Q and conduct an empirical study on five mathematics (Math-500, GSM8K and Omni-MATH) and coding (HumanEval+ and MBPP+) tasks. %taking differentiable ternarization as the baseline. 
Surprisingly, our empirical study shows that, even for the pre-trained Qwen3-4B~\citep{qwen3} with a twofold larger model scale, the resulting 1.58-bit model obtained via CAT-Q with learned group-wise scaling factors and weight thresholds performs significantly worse than BitNet b1.58 2B4T on these complex reasoning tasks, as illustrated by the gray-colored results in Figure~\ref{fig01:empirical_study}.%the absence of an effective calibration method causes 

We find that the performance collapse is caused by the calibration scheme. For pre-trained LLMs, %state-of-the-art 
existing PTQ methods mostly adopt sequential layer-wise or block-wise reconstruction, which relies on a set of calibration data to estimate the output distribution of each layer or block. Most commonly, calibration data is sampled from web text corpora such as C4~\citep{c4} and WikiText2~\citep{wikitext2}. However, calibration in this way has two limitations. First, it assumes that PTQ methods are robust to calibration variations. This %assumption 
tends to be valid under conventional low-bit formats (e.g., 8-bit/4-bit quantization), where their expressive capacities are reasonably sufficient to approximate the FP16 format, but it is less effective in the case of ternary quantization due to severe information loss, especially for complex reasoning tasks. Second, CoT demonstrations from pre-trained high-precision reasoning LLMs play an essential role in tackling mathematics and coding tasks, which break down a complex problem into a series of intermediate logical reasoning steps leading to a correct answer, whereas web text corpora mainly reflect generic local word-to-word relationships but not explicit logical reasoning patterns.

\textit{Motivated by this analysis, we present Attend to Your Own Thoughts (AYOT), a simple calibration method that serves as the core technical contribution of this paper}. AYOT combines the strengths of selecting appropriate calibration data and leveraging CoT demonstrations generated by the pre-trained high-precision target LLM. We apply AYOT to CAT-Q, forming a scalable ternary PTQ framework for reasoning LLMs, which we term ScaleQ-1.58. We conduct extensive experiments on five challenging mathematics and coding tasks used in the empirical study above, validating the desired scaling properties of ScaleQ-1.58 across diverse model architectures (including both dense and mixture-of-experts (MoE) models), varying model scales (from 1.7B to 235B parameters), and increasing amounts of calibration tokens (256K to 16M).\begin{figure}[htbp]
    \centering
    \vskip -0.05 in
    \centerline{\includegraphics[width=.9\textwidth]{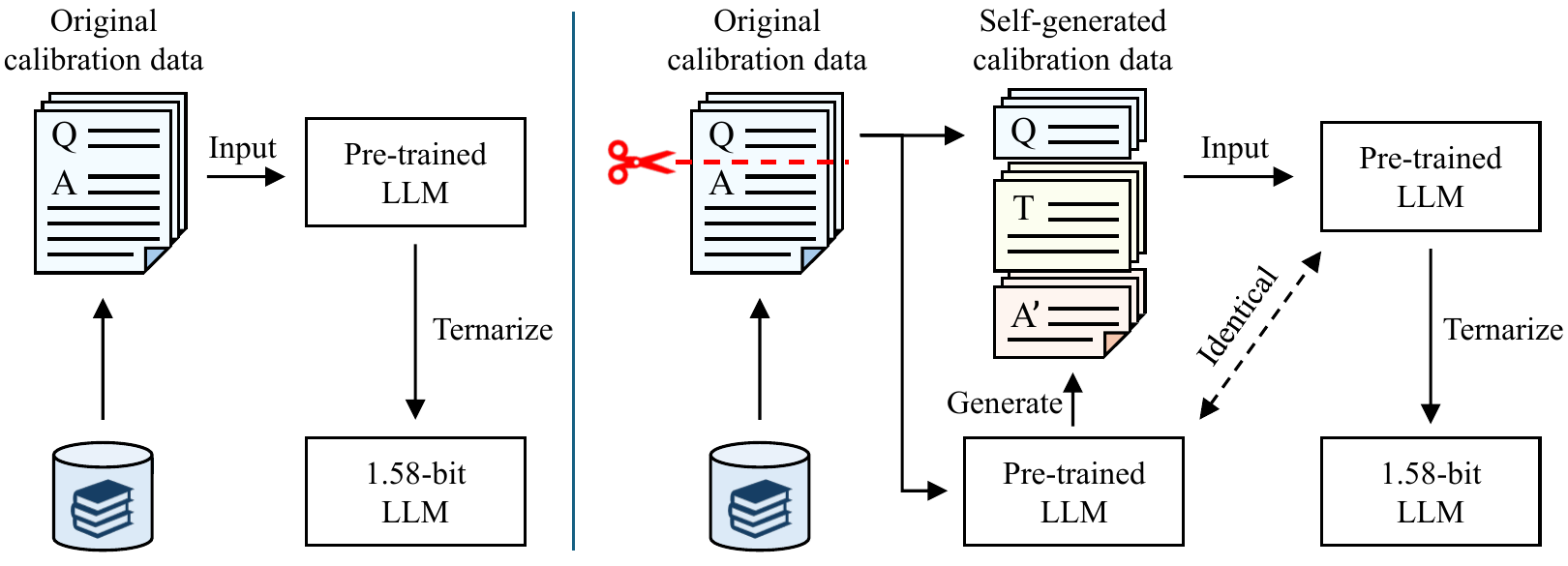}}
    \vspace{-0.05 in}
    \label{fig2:method}
    \caption{Post-training ternarization of LLMs under CoT-agnostic calibration (left) vs. CoT-aware calibration AYOT %Attend to Your Own Thoughts 
    (right). For the question Q, A denotes the ground-truth answer, and T and A' denote the reasoning traces and final answer generated by the pre-trained target LLM%to be ternarized
    , respectively.}
    \vskip -0.05 in
\end{figure}
In addition, we evaluate its favourable scalability across a broad range of tasks with varying difficulty levels, including mathematics, coding, scientific logic reasoning, commonsense reasoning, and basic language generation. Aligned with our goal, using only 4M calibration tokens, ScaleQ-1.58 takes 4 to 240 hours on a single server with 8 A100-80GB GPUs to produce diverse, state-of-the-art 1.58-bit reasoning LLMs (both dense and MoE) ranging from 1.7B to 235B parameters. In contrast, the prior best 1.58-bit reasoning LLM, BitNet b1.58 2B4T, is trained from scratch on 4T tokens, requiring 1,000,000$\times$ more tokens than our method. Intriguingly, AYOT also yields promising gains under other quantization bit-widths. % Our work makes scalable LLM ternarization technology accessible to the vast majority of researchers for the first time.
\section{Method}
In this section, we begin with a brief definition of CAT-Q~\citep{cat-q}, the fundamental module of ScaleQ-1.58, and then introduce %Attend to Your Own Thoughts 
AYOT, the core technical contribution of this work.

\subsection{Preliminary Concept}
% Recall that the determination of the scaling factor $\alpha$, the weight threshold $\Delta$, and gradients is essential in the optimization of ternary weight quantization. In the PTQ regime, this becomes even more challenging than in the QAT regime. As QAT can mitigate this issue, BitNet b1.58 2B4T~\citep{bitnet-2b4t} directly use statistical strategies to determine the scaling factor $\alpha$ and the weight threshold $\Delta$, and employ straight-through estimator (STE)~\citep{bnn} to approximate gradients during ternarization, following~\citep{bitnet-1.58, bitnetv2, spectra, tequila}. Unlike them, differentiable ternarization leverages a soft ternarization function:
Ternary % weight
quantization is known to be difficult to optimize owing to its extreme discreteness and non-differentiable nature. In the PTQ regime, only a limited set of calibration samples is available and retraining or fine-tuning is not permitted, making it significantly more challenging than in the QAT regime, which dominates current research on 1.58-bit LLMs~\citep{bitnet-1.58, bitnetv2, spectra, tequila, bitnet-2b4t}. To address this issue, CAT-Q introduces a soft ternarization function coupled with a learnable weight redistribution strategy:\begin{equation}
    %\small 
    %\resizebox{1\linewidth}{!}{
    T_i = f(\hat{W}_i;s,t,\Delta)=\frac{\tanh(ts(\hat{W}_i-\Delta))+\tanh(ts(\hat{W}_i+\Delta))} {2\tanh(ts)}.
    %}
   \label{eq2:dt}
   %\vspace{- 0.in}
\end{equation}Here, following the notation in Equation~\ref{eq1:twn}, $\hat{W}_i$ denotes the $i^{th}$ element of the redistributed pre-trained %high-precision 
weights $\hat{\textbf{W}}$ obtained by a linear transformation $\hat{\textbf{W}}=(\textbf{W}-\mu)/\alpha$ parameterized by two learnable variables: the weight mean $\mu$ and the scaling factor $\alpha$. Note that $\mu$ is used only for weight redistribution and is removed during weight reconstruction, leading to improved performance compared to using it in weight reconstitution while retaining the original hardware-friendly property of TWN~\citep{twn}. $s$ is a constant sharpness factor, and $t$ denotes the current calibration time-step normalized to $[0,1]$. %both of which 
The denominator $2\tanh(ts)$ controls the shape of the output curve, progressively bounding the output to $[-1,1]$ %over the course of 
during calibration. The weight threshold $\Delta$ determines the width of its zero-output region. In principle, with $f(\cdot)$, CAT-Q starts from the identity mapping $f(\cdot)=\hat{\textbf{W}}$ and asymptotically converges to a desired ternary output in a differentiable manner as $t$ increases. In implementation of CAT-Q, instead of the commonly used layer-wise weight reconstruction, it adopts cross-layer output reconstruction based on a sliding window~\citep{sliderquant}. Throughout this paper, for all experiments with ScaleQ-1.58, we use the same hyperparameter settings as in CAT-Q~\citep{cat-q}.

\subsection{Attend to Your Own Thoughts}

Existing PTQ methods typically rely on a small set of calibration samples to optimize quantization parameters. CAT-Q follows their de facto calibration practice by randomly selecting samples from web text corpora (e.g., C4~\citep{c4} and WikiText2~\citep{wikitext2}).\begin{figure}[htbp]
    \vspace{-0.05 in}
    \centering
    \includegraphics[width=0.9\columnwidth]{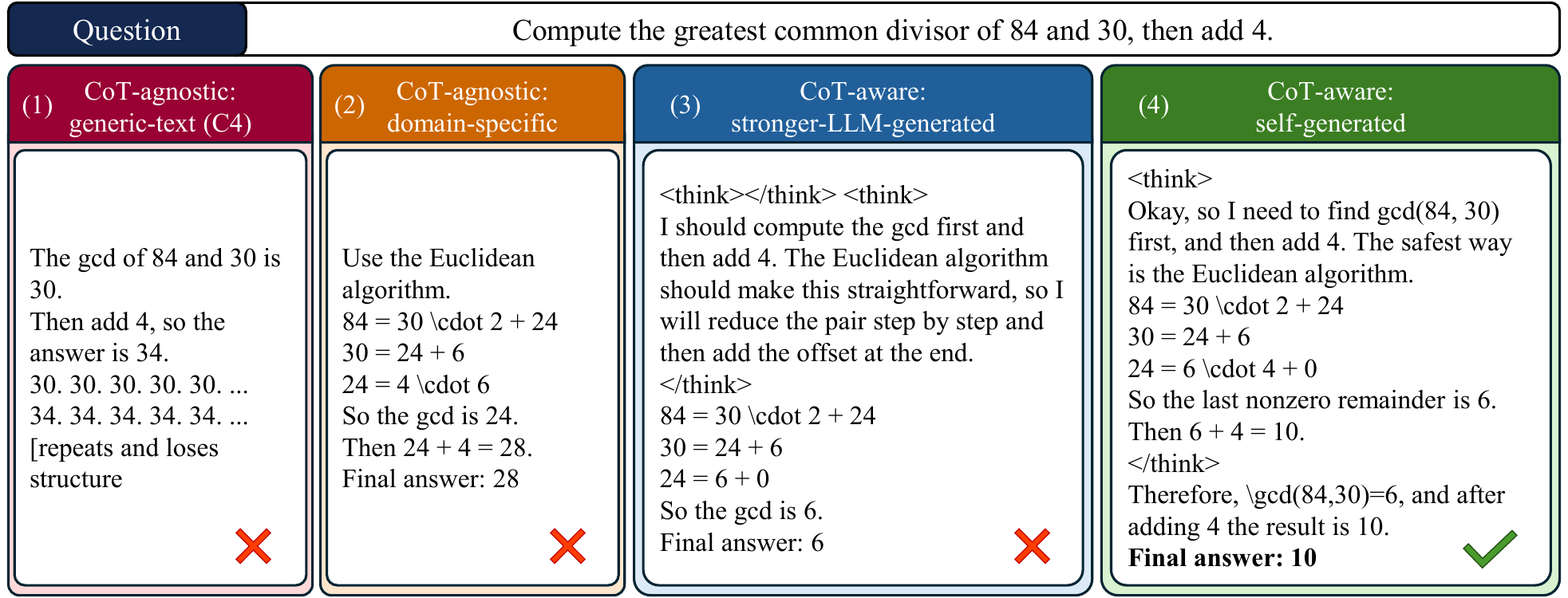}
    \vspace{-0.05 in}
    \caption{Reasoning examples of 1.58-bit Qwen3-4B models obtained via CAT-Q under different calibration schemes. Our AYOT significantly improves the capability of the resulting 1.58-bit model on complex mathematics tasks.~\textit{More examples are provided in the Appendix}.}
    \vspace{-0.05 in}
    \label{fig02:ayot_examples}
\end{figure} By introducing a new calibration method, Attend to Your Own Thoughts (AYOT) depicted in Figure~\ref{fig2:method}, ScaleQ-1.58 combines it with CAT-Q to achieve the basic goal of this paper: ternarizing pre-trained high-precision reasoning LLMs into 1.58-bit models that preserve the CoT reasoning ability across complex tasks.~\textit{AYOT is guided by two simple design principles: (1) selecting domain-specific calibration samples and (2) leveraging self-generated CoT demonstrations}, which are motivated by an empirical study exploring the scaling behavior of CAT-Q across five challenging mathematics (Math-500, GSM8K and Omni-MATH) and coding (HumanEval+ and MBPP+) tasks. In this study, using 4M calibration tokens throughout, we first test CAT-Q on pre-trained Qwen3-4B using the de facto calibration scheme, and observe severe performance collapse, resulting in zero or near-zero accuracy on most tasks, no matter calibration samples are drawn from C4 or WikiText2.
We conjecture that this catastrophic degradation %undesirable degeneration behavior
arises because web text corpora primarily encode generic word-to-word relationships and lack domain-specific content. Accordingly, we next use two public domain-specific datasets comprising question–answer pairs, MetaMathQA\citep{metamathqa} for mathematics and OpenCodeInstruct~\citep{opencodeinstruct} for coding, to randomly choose calibration samples evenly in each domain, having no sample overlaps with our five test task datasets. The average accuracy of the resulting 1.58-bit model across five tasks improves to 17.43\%, indicating the importance of selecting domain-specific calibration samples, namely the first design principle of AYOT, yet remains far from satisfactory. Considering that CoT interprets the step-by-step problem-solving process of pre-trained reasoning LLMs, we hypothesize that the lack of CoT demonstrations during ternarization may be the key reason for poor performance when domain-specific calibration data is used in a straightforward way, i.e., taking [\textit{question, answer}] pairs as the context input. %for ternary quantization. 
In light of this, we then employ two strategies to generate reasoning traces and final answer for each sampled question, forming [\textit{question, reasoning traces, generated answer}] triplets as the context input. Specifically, we first use DeepSeek-R1-671B~\citep{deepseek-r1}, %in FP8 format, 
a sufficiently large and powerful LLM, %as the teacher model, 
to generate CoT demonstrations for sampled questions, which serve as proxies for ground-truth reasoning traces and answers that are originally unavailable in MetaMathQA and OpenCodeInstruct. It further improves the mean accuracy from 17.43\% to 20.06\%, but it still remains low. This is likely because, under extreme ternary setting in the PTQ regime, the 1.58-bit model quantized from the pre-trained high-precision target LLM is unable to mimic the reasoning capability of a significantly larger and stronger LLM, due to differences in architecture and model scale. Therefore, the second design principle of AYOT is established: using the pre-trained target LLM itself to generate reasoning traces and answer for each sampled question. As a result, AYOT substantially boosts the mean accuracy to 45.60\%, yielding a remarkable absolute gain of 25.54\% over using DeepSeek-R1-671B. Compared to the prior best 1.58-bit reasoning LLM, BitNet b1.58 2B4T, trained from scratch on a massive amount of 4T tokens, our 1.58-bit Qwen3-4B produced by ScaleQ-1.58 on 4M calibration tokens shows an absolute gain of 8.97\%, while requiring 1,000,000$\times$ fewer tokens for quantization. 

Detailed results and illustrative reasoning examples of 1.58-bit Qwen3-4B models produced by CAT-Q under the above four calibration schemes are shown in Figure~\ref{fig01:empirical_study} and Figure~\ref{fig02:ayot_examples}, respectively. To better differentiate these four calibration schemes, we term (1) the first calibration scheme, the default setting of CAT-Q, as~\textit{CoT-agnostic: generic-text}; (2) the second calibration scheme, which uses domain-specific samples, as~\textit{CoT-agnostic: domain-specific}; (3) the third calibration scheme, which leverages CoT demonstrations and final answers generated by DeepSeek-R1-671B, as~\textit{CoT-aware: stronger-LLM-generated}; (4) AYOT, using the pre-trained high-precision target LLM itself to generate CoT demonstrations and final answers, as~\textit{CoT-aware: self-generated}.

\begin{table}[!t]
\centering
\setlength{\tabcolsep}{4pt} % 稍微增加了间距，因为列数变少了
\vskip -0.05 in
%\caption{Performance of different quantized LLMs on mathematical reasoning and code generation benchmarks.}
\caption{Performance of different LLMs ternarized by ScaleQ-1.58 on mathematics and coding tasks. We include the best prior 1.58-bit reasoning LLM, BitNet b1.58 2B4T (results from its paper~\citep{bitnet-2b4t}), for a comparison. The metric is accuracy (\%). \#Tokens denotes the number of tokens for quantization.}
\vskip -0.05 in
\renewcommand{\arraystretch}{1.2}
\resizebox{0.95\textwidth}{!}{
\begin{tabular}{c|c|c|ccc|cc}
\toprule
\textbf{Model} & \textbf{\#Bits} & \textbf{\#Tokens} & \textbf{Math-500~$\uparrow$} & \textbf{GSM8K~$\uparrow$} & \textbf{Omni-MATH~$\uparrow$} & \textbf{HumanEval+~$\uparrow$} & \textbf{MBPP+~$\uparrow$} \\
\midrule
BitNet b1.58 2B4T      & W1.58A16 & 4T             & 43.40 & 58.38 & - & 38.40 & 37.03  \\ \midrule     
\multirow{2}{*}{Qwen3-1.7B}  & W16A16 & -  & 90.80   & 84.08 & 27.76 & 76.22  & 53.97 \\
   ~    & W1.58A16 & \textbf{4M}    & 38.60  & 54.35  & 12.23  & 34.14  & 33.33    \\ \midrule  
% % \multirow{2}{*}{Qwen3.5-2B}  & W16A16 & -  & 80.20  &  71.27 & 27.24  & 39.02  & 29.07 \\
% %    ~    & W1.58A16 & \textbf{4M}    &  3.60   & 9.70  & 2.37  & 5.49  & 6.52   \\ \midrule  
\multirow{2}{*}{Qwen3-4B}  & W16A16 & -  & 96.80  & 88.10 & 34.64 & 85.37  & 61.90 \\
~      & W1.58A16 & \textbf{4M}    & 58.40   & 61.56   & 14.93 & 53.98    &   39.15   \\ \midrule 
\multirow{2}{*}{Qwen3-8B}  & W16A16 & -  &  96.80 & 88.85 & 34.67 & 87.20  & 63.49    \\
~      & W1.58A16 & \textbf{4M}    & 63.00 & 71.72 & 16.96 & 54.88  & 44.71  \\ \midrule
\multirow{2}{*}{Qwen3-14B}  & W16A16 & -  &  97.20   & 91.35 & 37.78 & 86.59  & 64.29      \\
~      & W1.58A16 & \textbf{4M}    &  76.80 &  79.19 & 22.49 & 74.39  & 50.26 \\ \midrule
\multirow{2}{*}{Qwen3-32B}  & W16A16 & -  &  96.80   & 93.63  & 44.08 & 87.20 & 66.93     \\
~     & W1.58A16 & \textbf{4M}    & 85.20  & 87.34  & 28.84 & 84.15   & 54.50  \\ \midrule
\multirow{2}{*}{Qwen3-30B-A3B}  & W16A16 & - &  97.60   & 94.84  & 41.10 & 86.59 & 62.96     \\
~ & W1.58A16 & \textbf{4M} & 76.80  & 84.99  & 23.62 & 60.37   & 44.97  \\  \midrule

\multirow{2}{*}{\makecell{DeepSeek-R1-\\Distill-Llama-70B}}  & W16A16 & -  &  96.60  & 93.03  & 44.26  & 88.41  & 61.90    \\
~     & W1.58A16 & \textbf{4M}    &  87.60  & 88.85  & 30.16  & 85.74  & 56.88 \\ \midrule

\multirow{2}{*}{Qwen3-235B-A22B}  & W16A16 & - & 98.60 & 96.28 & 45.51 & 89.02 & 69.31 \\
~ & W1.58A16 & \textbf{4M} & 78.80 & 85.97 & 26.88 & 67.07 & 48.51 \\

% \multirow{2}{*}{\makecell{Ring-flash-2.0 \\ (100B-A6.1B) }  } & W16A16 & - &  98.40   & 88.63  & 30.87 & 89.63 & 64.02     \\
% ~    & W1.58A16 & \textbf{4M}    & 36.20  & 76.48 & 14.55 &  34.12  & 32.23 \\ 
\bottomrule
\end{tabular}
}
\vskip -0.05 in

\label{tab:math_code_results}
\end{table}

\section{Experiments}

In this section, we extensively evaluate the efficacy of ScaleQ-1.58 across different LLM architectures %(including both dense and MoE) 
and model scales, as well as various quantization settings and downstream tasks. We also analyze the effect of key design choices, and~\textit{provide real deployment examples in the Appendix}.

\subsection{Setup} 

In our experiments, we focus on complex reasoning tasks including mathematics and coding, while also evaluating on scientific logic reasoning, basic language generation, and commonsense reasoning. Specifically, mathematics is evaluated on Math-500~\citep{math-500}, GSM8K~\citep{gsm8k} and Omni-MATH~\citep{omni-math}; coding is evaluated on HumanEval+ and MBPP+~\citep{evalplus}; scientific logic reasoning is evaluated on ProofWriter~\citep{proofwriter}. Basic language generation is measured by perplexity on WikiText2~\citep{wikitext2} and C4~\citep{c4}, and commonsense reasoning is reported as the average accuracy over PIQA~\citep{piqa}, ARC‑e, ARC‑c~\citep{arc}, HellaSwag~\citep{hellaswag}, and Winogrande~\citep{winogrande}. To validate the scaling properties of ScaleQ-1.58 across diverse model scales and architectures, we instantiate ScaleQ‑1.58 on the Qwen3 family, including dense models of 1.7B, 4B, 8B, 14B, and 32B parameters, as well as the MoE variants Qwen3‑30B‑A3B and Qwen3‑235B‑A22B. To assess generalization across different model families, we additionally include DeepSeek‑R1‑Distill‑Llama‑70B~\citep{deepseek-r1}, a strong distilled reasoning model. The calibration set is constructed by sampling prompts from MetaMathQA~\citep{metamathqa} and OpenCodeInstruct~\citep{opencodeinstruct}, generating responses with the pre-trained model to be quantized, and concatenating each prompt with its corresponding reasoning trace and answer to form context input. The total number of calibration tokens is 4M by default. \textit{Unless otherwise stated, we use W1.58A16 quantization.}

\subsection{Main Results}

\textbf{Scaling across Models and Tasks.} Table~\ref{tab:math_code_results} summarizes the ternary quantization results of ScaleQ-1.58 across a wide range of models (from 1.7B to 235B parameters, covering both dense and MoE architectures, and including Qwen and Llama families) on mathematics and coding tasks of varying difficulty levels, under the default budget of 4M calibration tokens. ScaleQ-1.58 performs consistently well across all models and tasks. We also find that smaller models are more sensitive to quantization than larger ones, and under comparable parameter counts, MoE models exhibit higher sensitivity than dense ones. Notably, compared to the best prior 1.58-bit reasoning LLM, BitNet b1.58 2B4T~\citep{bitnet-2b4t}, our ternary Qwen3-1.7B achieves competitive performance (over 90.52\% of the average score of BitNet b1.58 2B4T on four math/coding tasks) while using $1{,}000{,}000\times$ fewer tokens for quantization. Moreover, our ternary Qwen3-4B surpasses BitNet b1.58 2B4T by an absolute 8.97\%, highlighting the extreme efficiency of ScaleQ-1.58 in terms of the calibration token count for quantization.

\begin{figure}[t!]
    \vskip -0.05 in
    \centering
    \begin{minipage}[t]{0.85\textwidth}
        \centering
        \includegraphics[width=\textwidth]{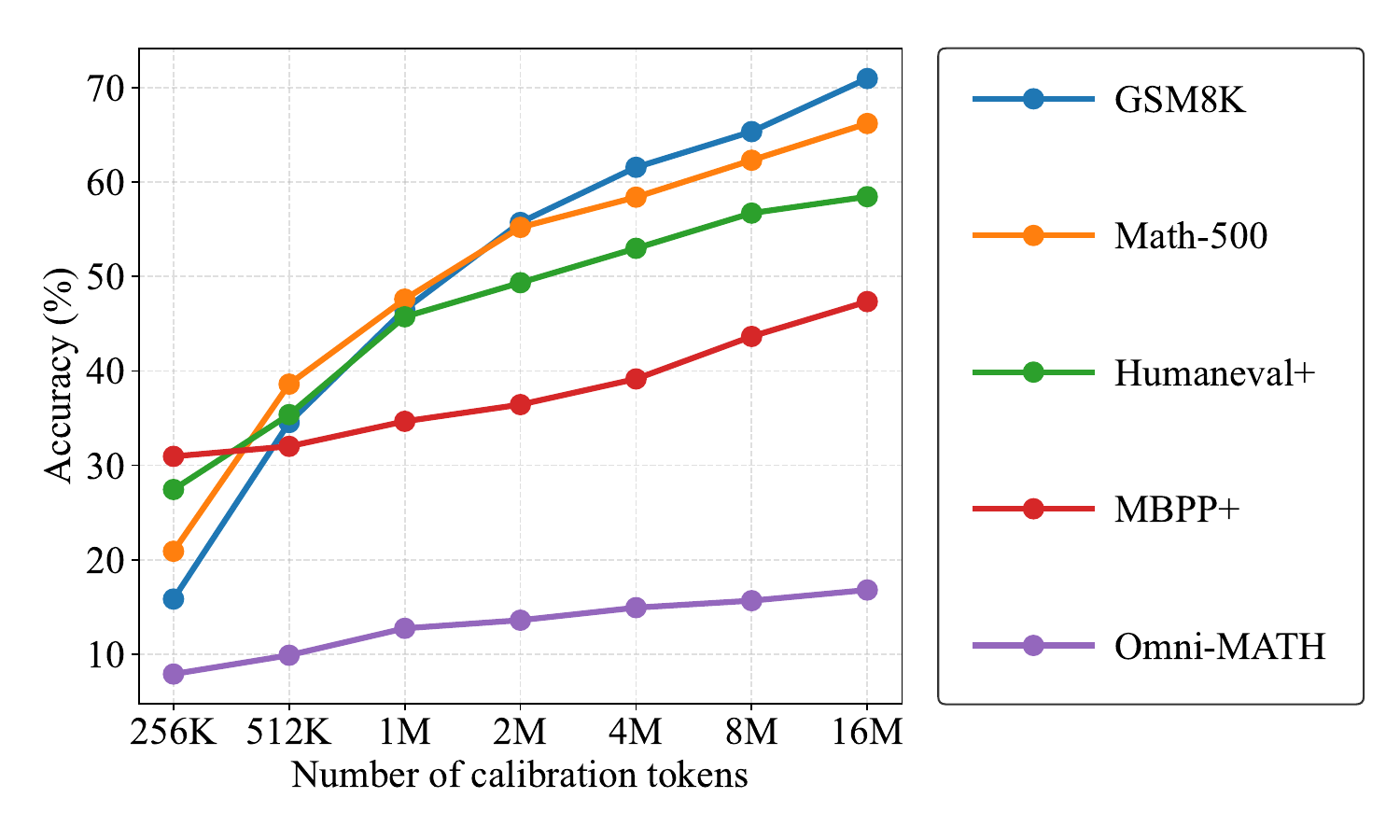}
    \end{minipage}
    \vskip -0.05 in
    \caption{Accuracy vs. number of calibration tokens for Qwen3-4B ternarized by ScaleQ-1.58.}
    \label{fig:scaling}
    \label{fig:scaling-log}
    \vskip -0.0 in
\end{figure}

\textbf{Scaling with Calibration Token Count.} 
We next examine the scaling behavior of ScaleQ-1.58 with respect to the calibration-token budget. As shown in Table~\ref{tab:scaling_results}, increasing the calibration token count from 256K to 16M tokens consistently improves the performance of the ternary Qwen3-4B across all evaluated tasks. Figure~\ref{fig:scaling} further shows that accuracy continues to increase with more calibration tokens, and the trend does not appear to plateau. This highlights the importance of the calibration token budget. Even with only 4M tokens our method already outperforms BitNet b1.58 2B4T, and as the calibration budget grows, this advantage becomes even more pronounced.

\begin{table}[!t]
\centering
\renewcommand{\arraystretch}{1.2}
%\caption{Performance scaling with quantization data size for Qwen3-4B quantized models on mathematics and coding tasks. Underlines denote the default configuration.}
\caption{Performance of Qwen3-4B ternarized by ScaleQ-1.58 with different numbers of calibration tokens on mathematics and coding tasks. Underlines denote our default configuration. All experiments use the same number of epochs and identical hyperparameters, isolating the effect of token budget.}
\vskip -0. in
\resizebox{0.95\textwidth}{!}{
\begin{tabular}{c|c|c|ccc|cc}
\toprule
\textbf{Model} & \textbf{\#Bits} & \textbf{\#Tokens} & \textbf{Math-500~$\uparrow$} & \textbf{GSM8K~$\uparrow$} & \textbf{Omni-MATH~$\uparrow$} & \textbf{HumanEval+~$\uparrow$} & \textbf{MBPP+~$\uparrow$} \\
\midrule
BitNet b1.58 2B4T  & W1.58A16 & 4T  & 43.40   & 58.38 & - & 38.40  & 37.03 \\ \midrule
\multirow{8}{*}{Qwen3-4B}      & W16A16 & -              & 96.80 & 88.10 & 34.64 & 85.37 & 61.90  \\ \cmidrule{2-8}   
~      & W1.58A16 & 256K    & 20.90   & 15.84  & 7.90 & 27.44    & 30.95     \\ 
~      & W1.58A16 & 512K    & 38.60   & 34.53  & 9.89 & 35.37    & 32.01     \\ 
~      & W1.58A16 & 1M    & 47.60   & 46.50  & 12.74 & 45.73    & 34.66     \\ 
~      & W1.58A16 & 2M    & 55.20   & 55.72  & 13.60 & 49.34    & 36.51     \\ 
~      & W1.58A16 & \underline{4M}    & \underline{58.40}   & \underline{61.56}  & \underline{14.93} & \underline{53.98}    & \underline{39.15}     \\ 
~    & W1.58A16 & 8M    & 62.31   & 65.34  & 15.67 & 56.71    & 43.65     \\ 
~      & W1.58A16 & 16M    & \textbf{66.20}   & \textbf{70.96}  & \textbf{16.80} & \textbf{58.45}    & \textbf{47.35}     \\ 
\bottomrule
\end{tabular}
}
\vskip -0.0 in
\label{tab:scaling_results}
\end{table}

\subsection{Ablation Studies}

We perform a lot of ablation studies on Qwen3-4B under W1.58A16 to identify the key factors for ScaleQ-1.58, choosing this model for its balanced size and decent performance. Additional experiments under W2A16 and W4A16 confirm the generality ability of our AYOT.

\begin{table}[!t]
\centering
\setlength{\tabcolsep}{3pt}
\vskip -0.0 in
\caption{Performance of Qwen3-4B ternarized by ScaleQ-1.58 with different calibration set sizes and training iterations on mathematics and coding tasks.}
\vskip -0.05 in
\renewcommand{\arraystretch}{1.2}
\resizebox{0.90\textwidth}{!}{
\begin{tabular}{c|c|c|ccc|cc}
\toprule
\textbf{\#Bits} & \textbf{\#Tokens} & \textbf{\#Iterations} & \textbf{Math-500~$\uparrow$} & \textbf{GSM8K~$\uparrow$} & \textbf{Omni-MATH~$\uparrow$} & \textbf{HumanEval+~$\uparrow$} & \textbf{MBPP+~$\uparrow$} \\
\midrule
W16A16 & -  & -  & 96.80 & 88.10 & 34.64 & 85.37 & 61.90  \\ \midrule     
%W1.58A16   & 256K  & 2560  & 20.80 & 15.85 & 7.90  & 27.44 & 30.95 \\
W1.58A16   & 256K  & 2560  & 20.90 & 15.84 & 7.90  & 27.44 & 30.95 \\
W1.58A16   & 256K  & 5120  & 27.40 & 22.74 & 8.99  & 29.27 & 31.75 \\  
W1.58A16   & 256K  & 10240 & 31.20 & 26.54 & 9.71  & 30.49 & 32.54 \\   \midrule

W1.58A16   & 2M & 2560  & 36.80 & 30.17 & 9.96 & 36.59 & 33.60 \\  
W1.58A16   & 2M & 5120  & 48.40 & 36.77 & 11.88 & 44.51 & 34.92 \\  
W1.58A16   & 2M & 10240 & \textbf{52.80} & \textbf{50.42} & \textbf{13.03} & \textbf{47.56} & \textbf{35.98} \\  
% W1.58A16   & 1024 & 20480 & 55.20 & 55.72 & 13.60 & 49.39 & 36.51 \\

\bottomrule
\end{tabular}
}
\vskip -0. in
\label{tab:training_config_results}
\end{table}

\begin{table}[!t]
\centering
\setlength{\tabcolsep}{4pt}
\vskip -0. in
\caption{Performance of Qwen3-4B ternarized by CAT-Q with different calibration schemes.}
\vskip -0.05 in
\renewcommand{\arraystretch}{1.2}
\resizebox{0.98\textwidth}{!}{
\begin{tabular}{l|ccc|cc}
\toprule
\textbf{Calibration Scheme} &  \textbf{Math-500~$\uparrow$} & \textbf{GSM8K~$\uparrow$} & \textbf{Omni-MATH~$\uparrow$} & \textbf{HumanEval+~$\uparrow$} & \textbf{MBPP+~$\uparrow$} \\
\midrule
CoT-agnostic: generic-text (Wikitext2)          & 0.00  & 2.65  & 0.43  & 0.00  & 0.00   \\
CoT-agnostic: generic-text (C4)        &  0.00  & 14.48  & 2.71  & 0.00  & 0.00  \\
CoT-agnostic: domain-specific          &  24.20  & 28.65  & 5.04  & 17.88  & 11.38    \\
CoT-aware: stronger-LLM-generated  & 31.80  & 30.78  & 6.98  & 16.46  & 14.29 \\
AYOT (CoT-aware: self-generated) &  \textbf{58.40}  & \textbf{61.56}  & \textbf{14.93}  & \textbf{53.98}  & \textbf{39.15}    \\ 
\bottomrule
\end{tabular}
}
\vskip -0.0 in
\label{tab:ablation_results}
\end{table}

\begin{table}[!t]
\centering
\setlength{\tabcolsep}{4pt}
\caption{Generalization ability of AYOT from W1.58A16 to other low-bit quantization settings. We integrate AYOT with SliderQuant~\citep{sliderquant} for W2A16 and W4A16 quantization on Qwen3‑4B.}
%\caption{Extended ablation study of calibration data for reasoning and code generation beyond the main \textbf{W1.58A16} setting, including \textbf{W2A16} and \textbf{W4A16} comparisons.  }
\vskip -0.05 in
\renewcommand{\arraystretch}{1.2}
\resizebox{0.9\textwidth}{!}{
\begin{tabular}{c|c|ccc|cc}
\toprule
\textbf{\#Bits} & \textbf{Calibration Schemes} &  \textbf{Math-500~$\uparrow$} & \textbf{GSM8K~$\uparrow$} & \textbf{Omni-MATH~$\uparrow$} & \textbf{HumanEval+~$\uparrow$} & \textbf{MBPP+~$\uparrow$} \\
\midrule
\multirow{2}{*}{W1.58A16} & CoT-agnostic: generic-text (C4)          &  0.00  & 14.48  & 2.71  & 0.00   & 0.00   \\
~ & AYOT+CAT-Q & \textbf{58.40} & \textbf{61.56} & \textbf{14.93} & \textbf{53.98} & \textbf{39.15} \\
\midrule
\multirow{2}{*}{W2A16} & CoT-agnostic: generic-text (C4)          &  1.40 &  16.48  & 3.39 & 1.22 & 0.79 \\
~ & AYOT+SliderQuant & \textbf{60.20} & \textbf{68.39} & \textbf{16.30} & \textbf{54.27} & \textbf{42.59} \\
\midrule
\multirow{2}{*}{W4A16} & CoT-agnostic: generic-text (C4) & 92.80 & 78.47 & 29.92 & 80.49 & 60.85 \\
~ & AYOT+SliderQuant & \textbf{95.00} & \textbf{83.02} & \textbf{33.33} & \textbf{83.54} & \textbf{63.49} \\
\bottomrule
\end{tabular}
}
\label{tab:ablation_results_2bit}
\end{table}

\textbf{Calibration Set Size vs. Quantization Budget.} 
Table~\ref{tab:scaling_results} shows that more calibration tokens improve the performance of ScaleQ-1.58 under the same number of calibration epochs. To dispel any concern that this gain might be due solely to more iterations, we further conduct a controlled analysis in Table~\ref{tab:training_config_results}. %With the same number of calibration samples, increasing optimization iterations yields consistent gains. However, the gains from scaling up calibration tokens are substantially more significant. 
With the same number of calibration tokens, increasing the number of optimization iterations yields consistent gains. For 256K tokens, raising the number of optimization iterations from 2560 to 10240 improves the average accuracy across the five tasks by 5.49\%. Yet under the same 10240 optimization iterations, scaling the calibration token count from 256K to 2M tokens produces a much larger gain of 13.86\% on average. This indicates that data diversity, not just the number of optimization iterations, drives the performance improvement.

\textbf{Different Calibration Schemes.} 
Having established the importance of calibration set size, we now study the impact of different calibration schemes. 
%we now turn to the role of calibration data content, specifically how the samples themselves are constructed. 
As a qualitative illustration, Figure~\ref{fig02:ayot_examples} presents how the ternary models respond with four different calibration schemes. In Table~\ref{tab:ablation_results}, we present a more detailed numerical comparison of their performance. Two main observations stand out. First, selecting domain-specific calibration samples leads to better performance than generic-text data (Wikitext2 and C4). Second, when using domain-specific data, self-generated calibration samples prove to be more effective than either the CoT-agnostic responses from the original dataset or CoT-aware responses generated by a stronger model. These results validate the two key principles proposed by our AYOT.

%  The gain over generic-text calibration is largest in the challenging 1.58-bit regime, yet our method remains consistently superior at higher bit-widths.
\textbf{Generalizing AYOT to Other Low-Bit Quantization Settings.} Next, we study the generalization ability of AYOT from ternary quantization to other low-bit quantization settings. As shown in Table~\ref{tab:ablation_results_2bit}, we integrate AYOT with SliderQuant~\citep{sliderquant} and test it under two typical PTQ settings (W2A16 and W4A16), besides testing ScaleQ1.58 under the extreme W1.58A16. We can see that our method consistently achieves superior performance under all bit-width setting compared to the baseline methods (CAT-Q and SliderQuant). Comparatively, the gains over generic-text calibration are especially pronounced in extremely low-bit quantization settings such as W1.58A16 and W2A16. This indicates that AYOT is not limited to ternary quantization, but can serve as a broadly effective calibration scheme across different bit-width settings.

\begin{table}[!t]
\centering
\setlength{\tabcolsep}{6pt}
\vskip -0.05 in
\caption{Effect of calibration sequence length and introducing task-specific calibration data for ScaleQ-1.58 on Qwen3-4B. When task-specific data is introduced, a part of the calibration data is replaced with samples from the training sets of GSM8K and MBPP while keeping total budget fixed at 4M tokens. Shaded cells denote target tasks aligned with the task-specific calibration data.}
\vskip -0.05 in
\renewcommand{\arraystretch}{1.2}
\resizebox{0.8\textwidth}{!}{
\begin{tabular}{c|c|ccc|cc}
\toprule
%\multirow{2}{*}{\textbf{Domain-Specific}} 
\multirow{2}{*}{\textbf{Task-Specific}} 
& \multirow{2}{*}{\textbf{SeqLen}} 
& \multicolumn{3}{c|}{\textbf{Mathematics}} 
& \multicolumn{2}{c}{\textbf{Coding}} \\
% \cmidrule(lr){4-6} \cmidrule(lr){7-8}
& 
& \textbf{Math-500$\uparrow$} & \textbf{GSM8K$\uparrow$} & \textbf{Omni-MATH$\uparrow$} 
& \textbf{HumanEval+$\uparrow$} & \textbf{MBPP+$\uparrow$} \\
\midrule
           & 2048 & 58.40 &  \cellcolor{blue!10}61.56 & 14.93 & 53.98 & \cellcolor{blue!10}39.15 \\
% \checkmark & \checkmark & 2048 & 58.40 & 69.37 & 15.46 & 54.54 & 41.35 \\
\checkmark & 2048 & 59.40 & \cellcolor{blue!10}70.45 & 15.21 & 55.37 & \cellcolor{blue!10}43.65 \\ \midrule
  & 4096 & 68.60 &  \cellcolor{blue!10}62.33 & 18.11 & 64.63 & \cellcolor{blue!10}42.06 \\
\checkmark      & 4096 & \textbf{68.80} &  \cellcolor{blue!10}\textbf{73.62} & \textbf{18.77} & \textbf{66.46} & \cellcolor{blue!10}\textbf{45.24} \\
\bottomrule
\end{tabular}
}
\label{tab:domain_task_ablation}
\vskip -0.0 in
\end{table}

\begin{table}[!t]
\centering
\setlength{\tabcolsep}{6pt}
\vskip -0.00 in
\caption{Generalization of ScaleQ-1.58 on Qwen3-4B to the scientific logic reasoning task ProofWriter. We compare calibration with and without a portion of its training samples.}
\vskip -0.05 in
\renewcommand{\arraystretch}{1.2}
\resizebox{0.9\textwidth}{!}{
\begin{tabular}{cc|ccccc}
\toprule
\multirow{2}{*}{\textbf{\#Bits}} &
\multirow{2}{*}{\textbf{Calibration Data Domains}} &
\multicolumn{5}{c}{\textbf{Scientific Logic Reasoning (ProofWriter)}} \\

& &
\textbf{Depth=1$\uparrow$} &
\textbf{Depth=2$\uparrow$} &
\textbf{Depth=3$\uparrow$} &
\textbf{Depth=5$\uparrow$} &
\textbf{Average$\uparrow$} 
\\
\midrule

W16A16 & -  & 92.83 & 89.33 & 84.83 & 72.67 & 84.92 \\
\midrule
W1.58A16 & Mathematics, Code &  47.33 & 42.33 & 36.00 & 32.33 & 39.50 \\

W1.58A16 & Mathematics, Code, ProofWriter &  91.33 & 85.83 & 80.00 & 71.67 & 82.21 \\

\bottomrule
\end{tabular}
}
\label{tab:proofwriter_calibration}
\skip -0. in
\end{table}

\begin{table}[!t]
\centering
\setlength{\tabcolsep}{6pt}
\vskip -0. in
\caption{Generalization of ScaleQ-1.58 on Qwen3-4B to a wide range of tasks. %via improving the diversity of calibration data. 
We consider four types of calibration data: Generic (C4), Math/Code (MetaMathQA, OpenCodeInstruct), Scientific (ProofWriter), and Task‑specific (GSM8K, MBPP). The table reports performance on mathematics \& coding (average accuracy of Math‑500, GSM8K, Omni-MATH, HumanEval+ and MBPP+), language modeling (average perplexity of WikiText2 and C4), commonsense reasoning (average accuracy of PIQA, ARC‑e, ARC‑c, HellaSwag and Winogrande), and scientific logic reasoning (ProofWriter).}
\skip -0.05 in
\renewcommand{\arraystretch}{1.2}
\resizebox{0.99\textwidth}{!}{
\begin{tabular}{cccc|ccccc|c|c|c}
\toprule

\multicolumn{4}{c|}{\textbf{Calibration Data Domains}} 
& \multicolumn{5}{c|}{\textbf{Mathematics \& Coding}} 
& \multirow{2}{*}{\makecell{\textbf{Language}\\\textbf{Modeling Avg.$\downarrow$}}}
& \multirow{2}{*}{\makecell{\textbf{Commonsense}\\\textbf{Reasoning Avg.$\uparrow$}}}
& \multirow{2}{*}{\makecell{\textbf{Scientific Logic}\\\textbf{Reasoning Avg.$\uparrow$}}} \\
Generic & Math/Code & Scientific & Task-Specific &
\textbf{Math-500$\uparrow$} & \textbf{GSM8K$\uparrow$} & \textbf{Omni-MATH$\uparrow$}
& \textbf{HumanEval+$\uparrow$} & \textbf{MBPP+$\uparrow$}
& & & \\
\midrule
\checkmark &   & & 
& 0.00 & 14.48 & 2.71 & 0.00 & 0.00  & \textbf{25.40} & \textbf{58.17}  & 1.38 \\
& \checkmark & & & 58.40 & 61.56 & 14.93 & 53.98 & 39.15  & 35.34  & 53.23 & 39.50 \\
\checkmark & \checkmark & & & 57.60  & 60.80  & 14.89 & 53.65 & 40.74 & 28.18  & 57.27  & 47.46  \\
\checkmark & \checkmark &  \checkmark & & 58.00  & 64.37  & 15.11  & 53.05  & 41.27 & 28.64  & 57.05  & \textbf{83.92}  \\
 
\checkmark & \checkmark &  \checkmark  &  \checkmark & \textbf{59.10} & \textbf{70.36}  & \textbf{16.71} & \textbf{54.27} & \textbf{42.06}  & 33.14 & 52.81 & 83.54 \\
\bottomrule
\end{tabular}
}
\label{tab:domain_generic_ablation}
\skip -0. in
\end{table}

\textbf{Task-Specific and Long-Context Calibration.} 
Table~\ref{tab:domain_task_ablation} examines two factors: task relevance of calibration data and sequence length. Replacing part of domain‑specific calibration data (general mathematics/code) with task‑specific samples from GSM8K and MBPP training sets substantially improves performance on the corresponding target tasks while slightly improving others. Extending the sequence length from 2048 to 4096 consistently boosts results across different tasks, with larger gains on long‑form reasoning benchmarks such as Math‑500 and HumanEval+. These results confirm that task‑specific data provides finer‑grained alignment beyond domain matching, and longer context captures more complete reasoning traces that better match the model’s own inference generation.

\textbf{Generalizing ScaleQ-1.58 to Scientific Logic Reasoning.} 
As shown in Table~\ref{tab:proofwriter_calibration}, ScaleQ‑1.58 generalizes effectively to the scientific logic reasoning task, ProofWriter, when using task‑specific calibration data (256 examples from the training set of ProofWriter). That means our method has good potential to generalize to other challenging reasoning tasks beyond mathematics and coding.

\textbf{Generalizing ScaleQ-1.58 to a Wide Range of Tasks via Diverse Calibration Data Composition.}
Table~\ref{tab:domain_generic_ablation} provides a comprehensive view of how different calibration compositions affect ternary model's performance on various downstream tasks including mathematics and coding, scientific logic reasoning, language modeling (perplexity) and commonsense reasoning. %, which are the typical focus of most existing PTQ methods. 
The calibration data choice strongly shapes downstream performance, data from each domain most benefits its corresponding tasks. Mixing data from multiple domains yields balanced performance across tasks. Finally, adding task‑specific training samples from the target benchmarks further improves the corresponding tasks.

\section{Related Work}

In the field of post-training compression research for LLMs, some prior studies have explored the impact of calibration data from different perspectives. Most existing methods~\citep{gptq,llm_pruner, slicegpt, sparsegpt, smoothquant, spqr, wanda, omniquant,awq,quarot} routinely use a small set of fixed-length samples for calibration, e.g., 128 or 256 samples randomly drawn from web text corpora such as C4 and WikiText2, each comprising 2048 tokens. % based on the assumption that the performance is robust to variations in calibration data. 
Williams and Aletras~\citep{williams2024_cal_ref1} present the first systematic study on the role of calibration data in post-training compression for commonsense reasoning tasks, considering two quantization methods for 4-bit weights, GPTQ~\citep{gptq} and SpQR~\citep{spqr}, as well as two pruning methods, SparseGPT~\citep{sparsegpt} and Wanda~\citep{wanda}. They show that the choice of calibration data sampled from different web text corpora or pre-training data will affect their performance. Similarly, many subsequent works also focus on commonsense reasoning tasks, but they investigate different aspects of calibration data for post-training compression.~\citep{self_calibration} evaluates GPTQ and AWQ~\citep{awq} for 4-bit weight quantization, together with SparseGPT and Wanda for pruning, using either real or synthetic data. Their results indicate that, for the tested two pruning methods, synthetic text generated by the target model itself usually yields higher accuracy than randomly sampled web text, which the authors attribute to its closer approximation of the pre-training data distribution. In contrast, they find that the tested two PTQ methods are much less sensitive to the choice of calibration data. The authors of~\citep{oh2025beyond_cal_ref9} study the impact of calibration sequence length using six post-training compression methods, including two PTQ methods SmoothQuant~\citep{smoothquant} and AWQ~\citep{awq}, and show that using shorter calibration sequences outperforms the commonly used fixed-length of 2048 tokens.~\citep{paglieri2024outliers_cal_ref6} finds that, newer LLMs such as Llama-2 and Llama3 are significantly less sensitive to varying calibration data than the older OPT model, when using GPTQ, AWQ, and SmoothQuant.~\citep{lee2023enhancing_cal_ref2} proposes a sequence-length-aware calibration strategy to improve PTQ in 4-bit weight and 8-bit activation quantization. RSQ~\citep{sung2025rsq_cal_ref8} reveals that, for fixed-length calibration data, a small subset of tokens, particularly those at the beginning and end of sequences, are more important for layer-wise quantization methods such as GPTQ~\citep{gptq} and QuaRot~\citep{quarot}. MBS~\citep{zeng2024multilingual_cal_ref5} introduces a calibration scheme tailored for improving LLM compression methods including GPTQ, whose core insight is to sample the calibration data of different languages proportionally to the language distribution of the model-specific pre-training data.
 
In summary, these prior studies primarily focus on conventional low-bit quantization settings (e.g., 4-bit weights or 4-bit weights and 8-bit activations) and relatively simple commonsense reasoning tasks. Recently,~\citep{benchmark_qrllm} benchmarks the impact of quantization on reasoning LLMs, considering five PTQ methods: GPTQ, AWQ, SmoothQuant, FlatQuant~\citep{flatquant} and KVQuant~\citep{kvquant} under 8-bit, 4-bit and 3-bit quantization settings. It reveals that these PTQ methods generally incur serious performance degradation when model precision is reduced to 3-bit. In addition, it finds that the domain of calibration data has little effect on the tested quantization methods, except for GPTQ. %, rather than extreme 1.58-bit quantization and more challenging reasoning tasks. 
In contrast, this work differs from them substantially in focus and method formulation.

\section{Conclusion}

In this paper, we present ScaleQ-1.58, a post-training ternarization framework for reasoning LLMs. Built upon CAT-Q, its core contribution is AYOT, a simple and effective calibration method, which enables the construction of 1.58-bit reasoning LLMs capable of handling complex reasoning tasks at substantially reduced cost, compared with existing state-of-the-art methods based on QAT. Through extensive experiments, we show that ScaleQ-1.58 exhibits favorable scaling properties across different LLM architectures, model scales, and calibration data sizes, as well as a wide range of tasks with varying difficulty levels. Besides, AYOT also generalizes well to other quantization bit widths. We hope ScaleQ-1.58 will inspire future research on building 1.58-bit reasoning LLMs in a cost-efficient and accurate manner.

\bibliography{neurips_2026}

%%%%%%%%%%%%%%%%%%%%%%%%%%%%%%%%%%%%%%%%%%%%%%%%%%%%%%%%%%%%

\newpage
\appendix

\section*{Supplementary Materials Contents }
\begin{itemize}[leftmargin=2em]
    \item Section~\ref{sec:limit}:  Limitations and broader impact.
    \item Section~\ref{sec:dataset}: Datasets used in experiments.
    \item Section~\ref{sec:implementation}: Implementation details of ScaleQ-1.58.
    \item Section~\ref{sec:more_results}: More experimental results.
    \item Section~\ref{sec:calibration_cost}: Ternarization time cost.
    \item Section~\ref{sec:real_quant}: Runtime acceleration on hardware.
    \item Section~\ref{sec:context_length}: Reasoning context length statistics.
    \item Section~\ref{app:qualitative-samples-v2}: Qualitative reasoning samples.
\end{itemize}

\setcounter{section}{0}
\setcounter{equation}{0}
\setcounter{figure}{0}
\setcounter{table}{0}

\renewcommand\thesection{\Alph{section}}
\renewcommand\thefigure{\Alph{figure}}
\renewcommand\thetable{\Alph{table}}
\renewcommand{\theequation}{\Alph{equation}}

% \begin{center}
% 	{
% 		\Large{\textbf{Appendix}}
% 	}
% \end{center}

\section{Limitations and Broader Impact} 
\label{sec:limit}
Although we have provided two real deployment cases of 1.58-bit reasoning LLMs in Table~\ref{tab:real_quant_benchmark} and shown clear inference advantages over standard 2-bit and 4-bit formats, the current deployments are still not fully optimized. In particular, public 1.58-bit GPU/CPU kernels implemented by the authors of BitNet b1.58 series mainly support its own model family, and efficient kernel support for more diverse 1.58-bit LLMs remains limited. In addition, while ScaleQ-1.58 is a scalable ternary PTQ method for pre-trained reasoning LLMs of diverse model architectures and scales, there is still considerable room to further reduce the performance gap to the FP16 baseline, especially on complex reasoning tasks that remain underexplored in LLM quantization research. That said, compared with BitNet b1.58 2B4T trained on 4T tokens, ScaleQ-1.58 already offers a substantially more practical trade-off between training cost and performance. For broader impact, our method can lower the deployment cost of pre-trained reasoning LLMs on edge, client and cloud devices, generating 1.58-bit high-performance reasoning LLMs in a cost-efficient PTQ manner and making them more accessible in real-world applications.

\section{Datasets Used in Experiments}
\label{sec:dataset}
\textbf{WikiText2}~\citep{wikitext2} is a popular language modeling benchmark consisting of over 2 million tokens from verified Wikipedia articles.

\textbf{C4}~\citep{c4}(Colossal Clean Crawled Corpus) is a large-scale dataset primarily used for language modeling tasks, comprising 156 billion clean tokens. It is sourced from cleaned web pages, originally from Common Crawl.

\textbf{PIQA}~\citep{piqa} contains 16,000 training and 3,084 test samples. It focuses on physical reasoning through multiple-choice questions, where models select the most appropriate solution from two options, with exactly one correct answer.

\textbf{ARC}~\citep{arc} is a dataset of 7,787 genuine grade-school level, multiple-choice science questions, assembled to encourage research in advanced question-answering. The dataset is partitioned into a Challenge Set and an Easy Set, where the former contains only questions answered incorrectly by both a retrieval-based algorithm and a word co-occurrence algorithm. 

\textbf{HellaSwag}~\citep{hellaswag} contains 70,000 training and 10,042 validation samples. It focuses on commonsense reasoning by predicting the most plausible sentence continuation, sourced from crowdsourced captions and activity descriptions.

\textbf{Winogrande}~\citep{winogrande} is a collection of 44,000 problems, which is formulated as a fill-in-a-blank task with binary options, the goal is to choose the right option for a given sentence which requires commonsense reasoning.

% \textbf{BoolQ}~\citep{boolq} is a dataset comprising 15,942 naturally occurring yes/no questions paired with Wikipedia passages. Each example consists of a question, a passage, and a binary answer, aiming to evaluate reading comprehension and entailment-like reasoning.

\textbf{MMLU}~\citep{mmlu} is a benchmark designed to evaluate the multitask accuracy of language models across 57 diverse subjects, including elementary mathematics, U.S. history, computer science, law, and more. The dataset consists of multiple-choice questions and is intended to assess models' world knowledge and problem-solving abilities in zero-shot and few-shot settings.

\textbf{MATH-500}~\citep{math-500} comprises 500 challenging competition-level mathematics problems sampled from the MATH dataset. These problems span various topics such as algebra, geometry, number theory, and probability, and are designed to test a model's ability to perform complex mathematical reasoning and generate step-by-step solutions.

\textbf{Omni-MATH}~\citep{omni-math} is a large-scale benchmark for olympiad-level mathematical reasoning, containing 4,428 problems collected from diverse national and international mathematics competitions. The dataset covers a broad range of advanced topics and is designed to evaluate models on difficult multi-step problem solving beyond standard school-level math benchmarks.

% \textbf{AIME-2024}~\citep{aime2024} includes 30 problems from the 2024 American Invitational Mathematics Examination (AIME), a prestigious high school mathematics competition. The dataset serves as a benchmark for evaluating models' capabilities in solving advanced mathematical problems that require deep understanding and creative problem-solving skills.

\textbf{GSM8K}~\citep{gsm8k} is a dataset of 8,792 high-quality, linguistically diverse grade school math word problems created by human problem writers. The dataset is segmented into 7,473 training problems and 1,319 test problems, each requiring multi-step reasoning and basic arithmetic operations to solve.

\textbf{HumanEval+}~\citep{evalplus} is an extension of the HumanEval dataset, consisting of 164 original programming problems designed to assess the functional correctness of code generated by language models. Each problem includes a function signature, a docstring specifying the intended functionality, and multiple test cases for evaluation.

\textbf{MBPP+}~\citep{evalplus} is an augmented version of the Mostly Basic Programming Problems (MBPP) dataset, comprising approximately 378 crowd-sourced Python programming tasks. Each task includes a natural language description, a reference solution, and three test cases, aiming to evaluate models' abilities in basic programming and problem-solving.

\textbf{ProofWriter}~\citep{proofwriter} is a scientific logic reasoning dataset built from synthetic natural language theories consisting of facts and rules, paired with hypotheses labeled as true, false, or unknown. It also provides accompanying proofs and varying reasoning depths, making it suitable for evaluating multi-step deductive inference.

\textbf{MetaMathQA}~\citep{metamathqa} is a large-scale mathematical instruction dataset constructed by bootstrapping and rewriting existing math problems into diverse question variants. It is designed to improve mathematical reasoning in language models and provides rich supervision spanning both grade-school and competition-style mathematics.

\textbf{OpenCodeInstruct}~\citep{opencodeinstruct} is a large-scale code instruction dataset released by NVIDIA, containing millions of instruction-response pairs for code generation and related programming tasks. The dataset also includes unit tests and quality annotations, making it useful for training or calibrating models on code-centric workloads.

\section{Implementation Details of ScaleQ-1.58}
\label{sec:implementation}

\subsection{Details of Calibration Data}

Our default calibration setup uses 2048 samples, each truncated or padded to 2048 tokens. For the default mathematics/coding configuration, we randomly sample 1024 examples from MetaMathQA and 1024 examples from OpenCodeInstruct, which serve as the mathematics-domain and code-domain calibration data, respectively. For the task-specific experiments in Table~\ref{tab:domain_task_ablation}, we replace one eighth of the mathematics-domain samples with examples from the GSM8K training set, and one eighth of the code-domain samples with examples from the MBPP training set, which does not overlap with the MBPP+ test set. When using ProofWriter as additional calibration data (shown in Table~\ref{tab:proofwriter_calibration}), we sample from the ProofWriter training set and replace one eighth of the mathematics-domain samples and one eighth of the code-domain samples accordingly. Unless otherwise stated, all calibration experiments are run for 60 epochs with a batch size of 3. Other implementation details follow CAT-Q~\citep{cat-q}.

\subsection{Quantization Details}
\label{sec:quant_details}

The weights are quantized in a group-wise manner with a fixed group size of $g=128$, and the ternary reconstruction uses $\mathbf{W}\approx\alpha\mathbf{T}$, 
discarding the weight mean $\mu$ learned to redistribute the pre-trained weights $\mathbf{W}$. As we have clarified in the Method section, this design retains the standard ternary format originally proposed in TWN and its hardware-friendly property. For MoE models, we leave the router layers in full precision and quantize the remaining modules.

\begin{table}[t!]
\centering
\setlength{\tabcolsep}{4pt}
\caption{Performance of 1.58-bit Qwen3-30B-A3B models produced by CAT-Q with different calibration schemes.}
\renewcommand{\arraystretch}{1.2}
\resizebox{0.8\textwidth}{!}{
\begin{tabular}{l|ccc|cc}
\toprule
\textbf{Calibration Schemes} & \textbf{Math-500~$\uparrow$} & \textbf{GSM8K~$\uparrow$} & \textbf{Omni-MATH~$\uparrow$} & \textbf{HumanEval+~$\uparrow$} & \textbf{MBPP+~$\uparrow$} \\
\midrule
CoT-agnostic: generic-text (C4) & 0.00 & 16.45 & 5.69 & 0.00 & 0.00 \\
AYOT (CoT-aware: self-generated) & \textbf{76.80} & \textbf{84.99} & \textbf{23.62} & \textbf{60.37} & \textbf{44.97} \\
\bottomrule
\end{tabular}
}
\label{tab:ablation_results_30ba3b}
\end{table}

\begin{table}[t]
\centering
\setlength{\tabcolsep}{4pt}
\caption{Detailed language modeling and commonsense reasoning results corresponding to Table~\ref{tab:domain_generic_ablation}.}
\renewcommand{\arraystretch}{1.2}
\resizebox{0.99\textwidth}{!}{
\begin{tabular}{ccccc|cc|cccccc}
\toprule
\multicolumn{5}{c|}{\textbf{Calibration Data Domains}} & \multicolumn{2}{c|}{\textbf{Language Modeling}} & \multicolumn{6}{c}{\textbf{Commonsense Reasoning}} \\
\cmidrule(lr){1-5} \cmidrule(lr){6-7} \cmidrule(lr){8-13}
\textbf{\#Bits} & \textbf{Generic} & \textbf{Math/Code} & \textbf{Scientific} & \textbf{Task-Specific}
& \textbf{WikiText2$\downarrow$} & \textbf{C4$\downarrow$}
& \textbf{PIQA$\uparrow$} & \textbf{ARC-e$\uparrow$} & \textbf{ARC-c$\uparrow$} & \textbf{HellaSwag$\uparrow$} & \textbf{Winogrande$\uparrow$} & \textbf{Avg.$\uparrow$} \\
\midrule
W16A16 &  &  &  &  & 13.67 & 16.53 & 75.19 & 78.32 & 53.84 & 68.41 & 65.51 & 68.25 \\
W1.58A16 & \checkmark &  &  &  & \textbf{26.40} & \textbf{24.40} & \textbf{71.44} & \textbf{64.19} & \textbf{38.93} & \textbf{55.26} & \textbf{61.04} & \textbf{58.17} \\
W1.58A16 &  & \checkmark &  &  & 39.23 & 31.45 & 64.17 & 58.22 & 37.48 & 49.64 & 56.62 & 53.23 \\
W1.58A16 & \checkmark & \checkmark &  &  & 29.74 & 26.61 & 68.77 & 63.83 & 38.23 & 54.79 & 60.75 & 57.27 \\
W1.58A16 & \checkmark & \checkmark & \checkmark &  & 30.10 & 27.17 & 70.61 & 63.87 & 36.58 & 54.54 & 59.67 & 57.05 \\
W1.58A16 & \checkmark & \checkmark & \checkmark & \checkmark & 37.38 & 28.89 & 66.59 & 54.76 & 34.39 & 48.15 & 60.14 & 52.81 \\
\bottomrule
\end{tabular}
}
\label{tab:domain_generic_ablation_detailed_lm_cs}
\end{table}

\begin{table}[t!]
\centering
\setlength{\tabcolsep}{6pt}
\caption{Detailed scientific logic reasoning (ProofWriter) results corresponding to Table~\ref{tab:domain_generic_ablation}.}
\renewcommand{\arraystretch}{1.2}
\resizebox{0.99\textwidth}{!}{
\begin{tabular}{ccccc|ccccc}
\toprule
\multicolumn{5}{c|}{\textbf{Calibration Data Domains}} & \multicolumn{5}{c}{\textbf{Scientific Logic Reasoning (ProofWriter)}} \\
\cmidrule(lr){1-5} \cmidrule(lr){6-10}
\textbf{\#Bits} & \textbf{Generic} & \textbf{Math/Code} & \textbf{Scientific} & \textbf{Task-Specific}
& \textbf{Depth=1$\uparrow$} & \textbf{Depth=2$\uparrow$} & \textbf{Depth=3$\uparrow$} & \textbf{Depth=5$\uparrow$} & \textbf{Average$\uparrow$} \\
\midrule
W16A16 &  &  &  &  & 92.83 & 89.33 & 84.83 & 72.67 & 84.92 \\
W1.58A16 & \checkmark &  &  &  & 1.67 & 1.17 & 1.33 & 1.33 & 1.38 \\
W1.58A16 &  & \checkmark &  &  & 47.33 & 42.33 & 36.00 & 32.33 & 39.50 \\
W1.58A16 & \checkmark & \checkmark &  &  & 57.67 & 51.67 & 42.50 & 38.00 & 47.46 \\
W1.58A16 & \checkmark & \checkmark & \checkmark &  & 92.50 & \textbf{88.67} & \textbf{80.33} & \textbf{74.17} & \textbf{83.92} \\
W1.58A16 & \checkmark & \checkmark & \checkmark & \checkmark & \textbf{92.83} & 87.17 & 80.00 & \textbf{74.17} & 83.54 \\
\bottomrule
\end{tabular}
}
\label{tab:domain_generic_ablation_detailed_pw}
\end{table}

\section{More Experimental Results}
\label{sec:more_results}

\subsection{Ablation on Calibration Schemes with Qwen3-30B-A3B}
\label{sec:ablation_30ba3b}

Building on Table~\ref{tab:ablation_results} in the main paper which reports results on Qwen3-4B, we further perform ablation of the calibration scheme on the larger Qwen3-30B-A3B model. As shown in Table~\ref{tab:ablation_results_30ba3b}, the generic-text calibration scheme again leads to model collapse on most mathematics and coding tasks, whereas AYOT remains effective on this larger MoE model.

%To complement Table~\ref{tab:ablation_results} in the main paper, we report the corresponding calibration-scheme ablation on Qwen3-30B-A3B. Table~\ref{tab:ablation_results_30ba3b} shows that the generic-text calibration scheme again collapses on most mathematics and coding tasks, whereas AYOT remains effective on this larger MoE model.

\subsection{Detailed Results of Calibration Data Composition}
\label{sec:domain_generic_detail}

Tables~\ref{tab:domain_generic_ablation_detailed_lm_cs} and~\ref{tab:domain_generic_ablation_detailed_pw} provide the detailed results of Table~\ref{tab:domain_generic_ablation} in the main paper. Specifically, they report the full results for language modeling, the individual commonsense reasoning results, and the per-depth  scientific logic reasoning (ProofWriter) results that are averaged in the main table.

\section{Ternarization Time Cost}
\label{sec:calibration_cost}

We also report the wall-clock cost of our ternarization procedure under the default settings used in the paper, namely 60 epochs and 4M calibration tokens. Table~\ref{tab:calibration_time_cost} summarizes the total ternarization time of ScaleQ-1.58 for different models, measured on a server with 8$\times$A100-80G GPUs.

\begin{table}[htbp]
\centering
\caption{Ternarization time cost of ScaleQ-1.58 under the default settings of 60 epochs and 4M calibration tokens measured on a server with 8×A100-80G GPUs.}
\label{tab:calibration_time_cost}
\resizebox{0.48\textwidth}{!}{
\begin{tabular}{l|c}
\toprule
\textbf{Model} & \textbf{Training Time (h)} \\
\midrule
Qwen3-1.7B & 4 \\
Qwen3-4B & 8 \\
Qwen3-8B & 20 \\
Qwen3-14B & 32 \\
Qwen3-32B & 65 \\
Qwen3-30B-A3B & 56 \\
DeepSeek-R1-Distill-Llama-70B & 128 \\
Qwen3-235B-A22B & 240 \\
\bottomrule
\end{tabular}
}
\end{table}

\section{Runtime Acceleration on Hardware}
\label{sec:real_quant}

We also evaluate the deployment benefits of 1.58-bit reasoning LLMs produced by ScaleQ-1.58, using \texttt{llama.cpp}. Table~\ref{tab:real_quant_benchmark} compares the deployed 1.58-bit Qwen3-4B model with standard 4-bit and 2-bit baselines. For the real deployment of our 1.58-bit Qwen3-4B model, we use the \texttt{TQ1\_0} implementation on CPU and the \texttt{TQ2\_0} implementation on GPU. The 1.58-bit model achieves the smallest memory footprint and the highest decoding throughput on both CPU and GPU. These results confirm that the proposed ternary PTQ method can generate high-performance 1.58-bit reasoning LLMs with practical inference gains.

\begin{table}[!ht]
\centering
\setlength{\tabcolsep}{5pt}
\caption{Real deployment cases of Qwen3-4B models in different low-bit formats with \texttt{llama.cpp}. We report profiling results on both CPU and GPU for standard 4-bit and 2-bit models, together with the 1.58-bit model obtained by ScaleQ-1.58, which is implemented as \texttt{TQ1\_0} on CPU and \texttt{TQ2\_0} on GPU. We set the batch size to 1 and the generation length to 512 tokens.}
%\caption{Real quantization benchmark results for Qwen3-4B deployed with \texttt{llama.cpp}. We report weight memory and decoding throughput on both CPU and GPU for standard 4-bit and 2-bit baselines, together with the real 1.58-bit deployment of ScaleQ-1.58, implemented as \texttt{TQ1\_0} on CPU and \texttt{TQ2\_0} on GPU. We set the batch size to 1 and generation length to 512. }
\renewcommand{\arraystretch}{1.2}
\resizebox{0.99\textwidth}{!}{
\begin{tabular}{c|c|cc|cc}
\toprule
\textbf{Model} & \textbf{Deployment Format} & \textbf{CPU Memory (GB)} & \textbf{CPU Throughput (Tokens/s)} & \textbf{GPU Memory (GB)} & \textbf{GPU Throughput (Tokens/s)} \\
\midrule
\multirow{3}{*}{Qwen3-4B} &  W4A16 (Q4\_K\_M) & 2.32 & 19.31 & 2.32 & 225.91 \\
 & W2A16 (Q2\_K\_M) & 1.55 & 23.84 & 1.55 & 268.51 \\
 & ScaleQ-1.58 (TQ1\_0 / TQ2\_0) & \textbf{1.01} & \textbf{37.35} & \textbf{1.17} & \textbf{320.34} \\
\bottomrule
\end{tabular}
}

\label{tab:real_quant_benchmark}
\end{table}

% \begin{table}[htbp]
% \centering
% \caption{Context length statistics for mathematical, coding, and scientific logic reasoning tasks.}
% \label{tab:dataset_stats}
% \small
% \begin{tabular}{l|r|rrr}
% \toprule
% Dataset & \#Samples & \multicolumn{3}{c}{Context Length} \\
% \cmidrule(lr){3-5}
%  &  & Min & Max & Avg. \\
% \midrule

% Math-500 & 500 & 1024 & 30045 & 5188.34 \\
% Omni-MATH & 4428 & 61 & 9611 & 7013.57 \\

% GSM8K & 1319 & 547 & 4576 & 1484.81 \\
% HumanEval+ & 164 & 406 & 32760 & 3358.81 \\
% MBPP+ & 378 & 372 & 32760 & 3277.46 \\

% ProofWriter D1 & 600 & 265 & 4421 & 1518.42 \\
% ProofWriter D2 & 600 & 388 & 4405 & 1883.96 \\
% ProofWriter D3 & 600 & 507 & 4432 & 2275.15 \\
% ProofWriter D5 & 600 & 569 & 4424 & 2674.88 \\
% ProofWriter (all depths) & 2400 & 265 & 4450 & 1967.83 \\
% \bottomrule
% \end{tabular}
% \end{table}

\section{Reasoning Context Length Statistics}
\label{sec:context_length}

We report dataset statistics to characterize the distributions of context lengths across different evaluation tasks. All statistics are computed on the corresponding test or validation sets, and context lengths are measured from the complete inference context produced by the high-precision Qwen3-4B model, including the question, the reasoning traces, and the generated answer. Table~\ref{tab:dataset_stats} summarizes the per-sample context lengths for mathematics, coding, and scientific logic reasoning tasks, while Table~\ref{tab:mc_context_stats} reports the corresponding statistics for commonsense reasoning tasks, which are typically formulated as multiple-choice problems and do not require generating long-form outputs. Complex reasoning tasks have much longer contexts than commonsense reasoning tasks.

\begin{table}[htbp]
\centering
\small

\begin{minipage}[t]{0.48\textwidth}
\centering
\caption{Context length statistics for mathematics, coding, and scientific logic reasoning tasks.}
\label{tab:dataset_stats}
\resizebox{\linewidth}{!}{
\begin{tabular}{l|c|ccc}
\toprule
\multirow{2}{*}{Dataset} & \multirow{2}{*}{\#Samples} & \multicolumn{3}{c}{Context Length} \\
\cmidrule(lr){3-5}
 &  & Min & Max & Avg. \\
\midrule
Math-500 & 500 & 1024 & 30045 & 5188.34 \\
Omni-MATH & 4428 & 61 & 9611 & 7013.57 \\
GSM8K & 1319 & 547 & 4576 & 1484.81 \\
HumanEval+ & 164 & 406 & 32760 & 3358.81 \\
MBPP+ & 378 & 372 & 32760 & 3277.46 \\
ProofWriter depth = 1 & 600 & 265 & 4421 & 1518.42 \\
ProofWriter depth = 2 & 600 & 388 & 4405 & 1883.96 \\
ProofWriter depth = 3 & 600 & 507 & 4432 & 2275.15 \\
ProofWriter depth = 5 & 600 & 569 & 4424 & 2674.88 \\
ProofWriter (all depths) & 2400 & 265 & 4432 & 1911.34 \\
\bottomrule
\end{tabular}
}
\end{minipage}
\hfill
\begin{minipage}[t]{0.38\textwidth}
\centering
\caption{Context length statistics for commonsense reasoning tasks.}
\label{tab:mc_context_stats}
\resizebox{\linewidth}{!}{
\begin{tabular}{l|c|ccc}
\toprule
\multirow{2}{*}{Dataset} & \multirow{2}{*}{\#Samples} & \multicolumn{3}{c}{Context Length} \\
\cmidrule(lr){3-5}
 &  & Min & Max & Avg. \\
\midrule
PIQA & 3084 & 9 & 244 & 35.43 \\
ARC-e & 2376 & 9 & 169 & 31.42 \\
ARC-c & 1172 & 10 & 179 & 36.63 \\
% BoolQ & 3270 & 33 & 1271 & 146.60 \\
HellaSwag & 10042 & 17 & 147 & 78.41 \\
Winogrande & 1267 & 16 & 37 & 21.97 \\
\bottomrule
\end{tabular}
}
\end{minipage}

\end{table}

% \begin{table}[htbp]
% \centering
% \caption{Context length statistics for commonsense reasoning tasks.}
% \label{tab:mc_context_stats}
% \small
% \begin{tabular}{l|r|rrr}
% \toprule
% Dataset & \#Samples & \multicolumn{3}{c}{Context Length} \\
% \cmidrule(lr){3-5}
%  &  & Min & Max & Avg. \\
% \midrule
% PIQA & 3084 & 9 & 244 & 35.43 \\
% ARC-Easy & 2376 & 9 & 169 & 31.42 \\
% ARC-Challenge & 1172 & 10 & 179 & 36.63 \\
% BoolQ & 3270 & 33 & 1271 & 146.60 \\
% HellaSwag & 10042 & 17 & 147 & 78.41 \\
% Winogrande & 1267 & 2 & 34 & 16.36 \\
% \bottomrule
% \end{tabular}
% \end{table}

%%%%%%%%%%%%%%%%%%%%%%%%%%%%%%%%%%%%%%%%%%%%%%%%%%%%%%%%%%%%
% \newpage

\section{Qualitative reasoning samples}
\label{app:qualitative-samples-v2}

Extending beyond the single example in Figure~\ref{fig02:ayot_examples} of the main paper, here we provide additional qualitative reasoning samples to further illustrate the effectiveness of our method. Specifically, we select four mathematics problems (as shown in Section~\ref{math_sample}) from Math-500 and four coding problems (as shown in Section~\ref{code_sample}) from HumanEval+. These examples highlight how different calibration schemes affect the CoT reasoning ability of the resulting 1.58-bit models. Using AYOT, the model produces more stable reasoning traces and more accurate final answers across both mathematics and coding tasks.

% qualitative_samples_v2.tex
% This file is intended to be \input{} into an existing paper appendix.
%
% Required in the main preamble:
% \usepackage{xcolor}
% \usepackage{listings}
% \usepackage[most]{tcolorbox}
% \tcbuselibrary{listings,skins,breakable,raster}
% \usepackage{amsmath,amssymb}

\definecolor{samplequestionbg}{RGB}{248,248,248}
%\definecolor{genericstrip}{RGB}{228,228,228}
\definecolor{genericstrip}{RGB}{254,218,219}
%\definecolor{directstrip}{RGB}{223,236,244}
\definecolor{directstrip}{RGB}{255,232,209}
%\definecolor{strongerstrip}{RGB}{247,233,214}
\definecolor{strongerstrip}{RGB}{230,240,250}
%\definecolor{selfstrip}{RGB}{225,242,228}
\definecolor{selfstrip}{RGB}{228,246,222}
\definecolor{samplequestion_colbacktitle}{RGB}{230,230,250}

\lstdefinestyle{samplelisting}{
  basicstyle=\ttfamily\scriptsize,
  columns=fullflexible,
  keepspaces=true,
  breaklines=true,
  breakatwhitespace=false,
  breakautoindent=false,
  breakindent=0pt,
  showstringspaces=false,
  aboveskip=0pt,
  belowskip=0pt
}

\newcommand{\samplerowheight}{0.31\textheight}
\newcommand{\setrowheight}[1]{\renewcommand{\samplerowheight}{#1}}
\newlength{\samplecolsep}
\setlength{\samplecolsep}{8pt}
\newcommand{\samplecolgap}{\hspace*{\samplecolsep}\ignorespaces}

\newtcblisting{samplequestion}{
  listing only,
  enhanced,
  width=0.987\linewidth,
  center,
  colback=white,
  colframe=black!35,
  %colbacktitle=black!6,
  colbacktitle=samplequestion_colbacktitle,
  coltitle=black,
  title={Question},
  fonttitle=\bfseries\footnotesize,
  boxrule=0.45pt,
  arc=1.5pt,
  left=1mm,
  right=1mm,
  top=0.8mm,
  bottom=0.8mm,
  listing options={style=samplelisting}
}

\newtcblisting{samplebox}[2][]{
  listing only,
  enhanced,
  width=\linewidth,
  colback=white,
  colframe=black!35,
  coltitle=black,
  title={#2},
  fonttitle=\bfseries\footnotesize,
  boxrule=0.45pt,
  arc=1.5pt,
  left=1mm,
  right=1mm,
  top=0.8mm,
  bottom=0.8mm,
  height=\samplerowheight,
  valign=top,
  listing options={style=samplelisting},
  #1
}

\newenvironment{samplegrid}
  {\par\noindent\ignorespaces}
  {\par}

\newcommand{\samplegridsep}{\par\medskip\noindent\ignorespaces}

\newenvironment{samplecol}
  {\begin{minipage}[t]{\dimexpr(\linewidth-\samplecolsep)/2-3pt\relax}}
  {\end{minipage}}

\newcommand{\samplecase}[2]{%
  \subsubsection*{#1}%
}

\newcommand{\sampleobs}[1]{%
  \par\smallskip\noindent\textbf{Takeaway.} #1\par\medskip
}

% We complement the quantitative ablation with a qualitative comparison over the
% same curated examples. These excerpts are meant to visualize the concrete
% failure modes induced by different calibration settings: repetitive arithmetic,
% malformed closing, incorrect late-stage correction, unfinished code, and
% code-like outputs that still miss the task semantics. Across both math and
% code, the CoT-aware Self-generated setting is designed to look like the most
% stable and most Qwen3-like reasoning mode.

% The math cases show several distinct degradation patterns. The CoT-agnostic
% generic-text setting often fails before the model even enters a coherent
% derivation; the CoT-agnostic Domain-specific setting can carry the reasoning
% for several steps and still collapse at the end; the CoT-aware:
% teacher-generated setting often preserves a plausible scaffold but still has
% unstable endings. The CoT-aware Self-generated setting is consistently the
% cleanest: the chain-of-thought is longer, more coherent, and more faithful to
% the expected Qwen3 style.
\newpage

\subsection{Mathematics Reasoning}
\label{math_sample}

%\samplecase{Divisibility Confusion on a Simple GCD Task}{math-500\_39}

\begin{samplequestion}
What is the result when the greatest common factor of 6432 and 132 is increased by 11?
\end{samplequestion}

\begin{samplegrid}
\setrowheight{0.29\textheight}
\begin{samplecol}
\begin{samplebox}[colbacktitle=genericstrip]{CoT-agnostic: generic-text (C4)}
1. The greatest common factor (GCF) of 6432 and 132 is 132. So, the GCF is 132. Then, adding 11 to that GCF would be 143. But that's not the GCF.

2. The GCF of 6432 and 132 is 132. So adding 11 to 132 is 143, but that's not the GCF. So the answer is still 132.

3. The GCF of 6432 and 132 is 132. So adding 11 to 132 is 143, but that's not the GCF. So the answer is still 132.

4. The GCF is 132, and increasing it by 11 gives 143, but the GCF is 132. So the answer is 132.

...
332. 332. 332. 332. 332. 332. 332. ...
[truncated repetitive content]
\end{samplebox}
\end{samplecol}\samplecolgap
\begin{samplecol}
\begin{samplebox}[colbacktitle=directstrip]{CoT-agnostic: domain-specific}
...
Okay, let's solve this problem step by step.

First, we need to find the greatest common factor (GCF) of 6432 and 132. To find the GCF, we can use the Euclidean algorithm.

Step 1: Divide 6432 by 132.
6432 / 132 = 49.0... (since 132 x 49 = 6432). So the remainder is 0. That means that 132 is a factor of 6432, and the GCF is 132.

Wait, that's not correct. Let me check again. Wait, 132 x 49 = 6432. So the remainder is 0, which means that 132 is a factor of 6432. Therefore, the GCF of 6432 and 132 is 132.

Now, we need to add 11 to the GCF.
132 + 11 = 143.

The answer is $\boxed{143$.

\end{samplebox}
\end{samplecol}
\samplegridsep
\setrowheight{0.49\textheight}
\begin{samplecol}
\begin{samplebox}[colbacktitle=strongerstrip]{CoT-aware: stronger-LLM-generated}
<think>
Let me first compute the greatest common factor and only then adjust the result.
This should be easy, so I only need to be careful not to skip the divisibility step.
</think>

We need the greatest common factor of 6432 and 132, then we add 11.

Since 132 goes into 6432 exactly, the gcd is 132.

Indeed,

$$
6432 \div 132 = 48
$$

with remainder 0, so the common factor is 132.

Then

$$
132 + 11 = 143
$$

So the result is

$$
\boxed{143}
$$

The greatest common factor of 6432 and 132 increased by 11 is $\boxed{143}$.
\end{samplebox}
\end{samplecol}\samplecolgap
\begin{samplecol}
\begin{samplebox}[colbacktitle=selfstrip]{CoT-aware: self-generated}
<think>
Okay, so I need to find the greatest common factor of 6432 and 132, and then add 11. Let me slow down and do the gcd carefully, because this is exactly the kind of place where it is easy to accidentally confuse ``132 divides 6432'' with ``132 is the gcd.''

The cleanest way is the Euclidean algorithm.

First,
6432 = 132 \cdot 48 + 96
because 132 \cdot 48 = 6336, and 6432 - 6336 = 96.

So
\gcd(6432,132) = \gcd(132,96).

Now continue:
132 = 96 + 36,
96 = 2 \cdot 36 + 24,
36 = 24 + 12,
24 = 2 \cdot 12 + 0.

So the last nonzero remainder is 12, which means the gcd is 12.
Then 12 + 11 = 23.
</think>

Using the Euclidean algorithm:
1. $\gcd(6432,132)=\gcd(132,96)$
2. $\gcd(132,96)=\gcd(96,36)$
3. $\gcd(96,36)=\gcd(36,24)$
4. $\gcd(36,24)=\gcd(24,12)$
5. $\gcd(24,12)=12$

Final Answer: $\boxed{23}$
\end{samplebox}
\end{samplecol}
\end{samplegrid}
\captionof{figure}{Reasoning examples of 1.58-bit Qwen3-4B models obtained via differentiable ternarization under different calibration schemes. The mathematics problem involves calculating the greatest common factor.} %Our AYOT significantly improves the capability of the resulting 1.58-bit model on complex mathematics tasks.}

\newpage

%\sampleobs{The contrast here is not just ``right versus wrong.'' The two CoT-agnostic settings and the CoT-aware Stronger-LLM-generated setting all lock onto the same incorrect divisor pattern, while the CoT-aware Self-generated setting sustains a longer and more faithful reasoning chain before closing cleanly at the correct answer.}

%\samplecase{Correct Derivation, Wrong Final Value}{math-500\_93}

\begin{samplequestion}
For what value of $c$ will the circle with equation $x^2 - 10x + y^2 + 6y + c = 0$ have a radius of length 1?
\end{samplequestion}

\begin{samplegrid}
\setrowheight{0.35\textheight}
\begin{samplecol}
\begin{samplebox}[colbacktitle=genericstrip]{CoT-agnostic: generic-text (C4)}
</think>
</think>
Okay, let's try to figure out this problem. The equation given is x^2 - 10x + y^2 + 6y + c = 0.
We need to find the value of c such that the circle has a radius of 1.

First, I need to understand what this equation represents. It's a circle equation, but it's not the standard one.
The standard equation for a circle is x^2 + y^2 + c = 0, but this one has more terms.

Wait, maybe this equation is an equation of a circle, but with some extra terms.
Let me think about how equations like this can be used to describe a circle.

Maybe I should move terms around first. Or maybe I should compare it directly with the radius formula.
...
The. The answer. The. The. The. The. ...
[truncated repetitive content]
\end{samplebox}
\end{samplecol}\samplecolgap
\begin{samplecol}
\begin{samplebox}[colbacktitle=directstrip]{CoT-agnostic: domain-specific}
<think></think> <think> 
I should convert this quadratic equation into the standard form of a circle. Completing the square in both variables will expose the radius term directly.

We are given $x^2 - 10x + y^2 + 6y + c = 0.$

Completing the square gives $ (x-5)^2 + (y+3)^2 = 34 - c. $

For a circle of radius 1, the right-hand side should be 1, so the intended value is 33.

But let me restate the transformed equation once more:

$$
(x-5)^2 + (y+3)^2 = 34
$$

...
33
33
33
[truncated repetitive content]

\end{samplebox}
\end{samplecol}
\samplegridsep
\setrowheight{0.43\textheight}
\begin{samplecol}
\begin{samplebox}[colbacktitle=strongerstrip]{CoT-aware: stronger-LLM-generated}
<think>
Okay, let's solve the problem step by step.

First, we need to find the value of $c$ such that
$x^2 - 10x + y^2 + 6y + c = 0$ has a radius of 1.

For the $x$ terms:
$$x^2 - 10x = (x - 5)^2 - 25$$

For the $y$ terms:
$$y^2 + 6y = (y + 3)^2 - 9$$

So
$$(x - 5)^2 + (y + 3)^2 = 34 - c.$$

Since the radius is 1:
$$34 - c = 1,$$ so $c = 33$.

Let me check the steps again...
Actually, after checking, I should make sure whether the constant is 34 or 33 in the final answer box.
</think>

Wait, I must check the answer. The answer is $\boxed{34}$.
[truncated repetitive content]
\end{samplebox}
\end{samplecol}\samplecolgap
\begin{samplecol}
\begin{samplebox}[colbacktitle=selfstrip]{CoT-aware: self-generated}
<think>
Okay, so I need to determine $c$ so that the circle has radius 1. The main step is clearly to complete the square in $x$ and $y$ and compare with the standard circle form.

Starting from

$x^2 - 10x + y^2 + 6y + c = 0$.

For the x-part:
$x^2 - 10x = (x - 5)^2 - 25$.

For the y-part:
$y^2 + 6y = (y + 3)^2 - 9$.

Substituting those back in gives

$(x - 5)^2 + (y + 3)^2 = 34 - c$.

Now compare this with $(x-h)^2 + (y-k)^2 = r^2$.
Since the radius is 1, we need $r^2 = 1$, so $34 - c = 1$.
That gives $c = 33$.
</think>

Thus,
$$34 - c = 1$$
so
$$c = 33.$$

Final Answer: $\boxed{33}$
\end{samplebox}
\end{samplecol}
\end{samplegrid}
\captionof{figure}{Reasoning examples of 1.58-bit Qwen3-4B models obtained via differentiable ternarization under different calibration schemes. The mathematics problem involves solving for an unknown in a circle equation.}

\newpage

%\sampleobs{The CoT-aware Stronger-LLM-generated setting already contains the correct derivation, but it still loses control at the very end. The CoT-agnostic Domain-specific setting also reaches the correct answer but is unable to halt the repetitive output.}

\newpage

%\samplecase{Modular Inverse Misread}{math-500\_249}

\begin{samplequestion}
Find the unique odd integer $t$ such that $0<t<23$ and $t+2$ is the inverse of $t$ modulo $23$.
\end{samplequestion}

\begin{samplegrid}
\setrowheight{0.35\textheight}
\begin{samplecol}
\begin{samplebox}[colbacktitle=genericstrip]{CoT-agnostic: generic-text (C4)}
</think>
Okay, so I need to find an odd integer t that is less than 23 and such that t + 2 is the inverse of t modulo 23.

The question is asking for an odd integer t that is less than 23, and when you add 2 to it, the result is the inverse of t modulo 23.

Wait, but how does that work? Let me think.
If the inverse of t is the number that when you add 2 to it, it's the inverse of t...
Maybe the inverse is the number that when you add 2 to it, the result is the inverse of t.

For example, if t is 1, then the inverse would be 21, but that's not really the inverse.
Maybe I should try small odd values and see what happens.

...
the to the to to the to to the to to the to ...
[truncated repetitive content]
\end{samplebox}
\end{samplecol}\samplecolgap
\begin{samplecol}
\begin{samplebox}[colbacktitle=directstrip]{CoT-agnostic: domain-specific}
...
Okay, let's solve the problem step by step.

We need to find an odd integer $t$ such that $0 < t < 23$ and $t + 2$ is the inverse of $t$ modulo 23.

This means
$$(t + 2) \cdot t \equiv 1 \mod 23.$$

So
$$t^2 + 2t - 1 \equiv 0 \mod 23.$$

Now, we can check odd integers between 1 and 21.

1. $t = 1$ gives 2
2. $t = 3$ gives 14
3. $t = 5$ gives 11
4. $t = 7$ gives 16
5. $t = 9$ gives 6
6. $t = 11$ gives 4
...

So after checking, the answer should be 15.

So, the answer is $\boxed{15$.

\end{samplebox}
\end{samplecol}
\samplegridsep
\setrowheight{0.45\textheight}
\begin{samplecol}
\begin{samplebox}[colbacktitle=strongerstrip]{CoT-aware: stronger-LLM-generated}
<think>
Let me translate the inverse statement into a modular equation first.
</think>

<think></think>

The condition says that $t+2$ is the inverse of $t$ modulo 23.

So I interpret that as

$$
t + 2 \equiv 1 \pmod{23}.
$$

Then

$$
t \equiv -1 \pmod{23},
$$

which means

$$
t \equiv 22 \pmod{23}.
$$

So the only possible value in the range is 22, but that is even.

...
\mod 23 \mod 23 \mod 23 \mod 23 ...
[truncated repetitive content]
\end{samplebox}
\end{samplecol}\samplecolgap
\begin{samplecol}
\begin{samplebox}[colbacktitle=selfstrip]{CoT-aware: self-generated}
<think>
Okay, so I need to find an odd integer $t$ with $0 < t < 23$ such that $t + 2$ is the multiplicative inverse of $t$ modulo 23. So the first thing to do is translate that sentence into an equation modulo 23.

If $t + 2$ is the inverse of $t$, then their product should be congruent to 1 modulo 23:

$t(t+2) \equiv 1 \pmod{23}$.

That expands to

$t^2 + 2t - 1 \equiv 0 \pmod{23}$.

Now I can solve this quadratic modulo 23. Formally,

$t = \frac{-2 \pm \sqrt{8}}{2}$.

Since $10^2 \equiv 8 \pmod{23}$, the two solutions are $t = 4$ and $t = 17$.
Among these, only 17 is odd and satisfies the range constraint.
</think>

Therefore the unique odd integer is

$$
\boxed{17}
$$
\end{samplebox}
\end{samplecol}
\end{samplegrid}
\captionof{figure}{Reasoning examples of 1.58-bit Qwen3-4B models obtained via differentiable ternarization under different calibration schemes. The mathematics problem involves solving a modular inverse equation.}

\newpage

%\sampleobs{The four calibration schemes fail in clearly different ways here: the CoT-agnostic Generic-text setting never enters a useful modular reasoning path, the CoT-agnostic Domain-specific setting writes the right equation but closes with the wrong value, and the CoT-aware Stronger-LLM-generated setting misreads the inverse condition. Only the CoT-aware Self-generated setting sustains a complete modular reasoning chain.}

%\samplecase{Right Midpoint, Corrupted Ending}{math-500\_305}

\begin{samplequestion}
If $\frac{\sqrt{2x}}{\sqrt{3x-1}}=\frac32$, solve for $x$. Express your answer in simplest fractional form.
\end{samplequestion}

\begin{samplegrid}
\setrowheight{0.33\textheight}
\begin{samplecol}
\begin{samplebox}[colbacktitle=genericstrip]{CoT-agnostic: generic-text (C4)}
</think>
</think>
Okay, so I need to solve this problem where the ratio of sqrt(2x) over sqrt(3x-1) equals 3/2.

First, let me understand what the problem is asking.
The equation is sqrt(2x)/sqrt(3x-1) = 3/2.

So, maybe I need to think about the ratio as being 2x over 3x-1, but then multiplied by...
Wait, maybe that's not the right way to think about it.

Let me think about the equation again.
Maybe if I square it directly, something simple happens.
Or maybe I should isolate one square root first.

...
Let. Let. Let. Let. Let. Let. ...
[truncated repetitive content]
\end{samplebox}
\end{samplecol}\samplecolgap
\begin{samplecol}
\begin{samplebox}[colbacktitle=directstrip]{CoT-agnostic: domain-specific}
To solve the equation
$\frac{\sqrt{2x}}{\sqrt{3x - 1}} = \frac{3}{2}$, we follow these steps.

Step 1: Cross-multiply
$$2\sqrt{2x} = 3\sqrt{3x - 1}$$

Step 2: Square both sides
$$8x = 27x - 9$$

Step 3: Solve
$$x = \frac{9}{19}$$

Step 4: Check the solution
Substituting back gives the correct ratio.

So the solution is 9/19.
Let me write the final answer.

...

The answer is $\boxed{99}$.
The answer is $\boxed{99}$.
[truncated repetitive content]
\end{samplebox}
\end{samplecol}
\samplegridsep
\setrowheight{0.43\textheight}
\begin{samplecol}
\begin{samplebox}[colbacktitle=strongerstrip]{CoT-aware: stronger-LLM-generated}
<think>
I can remove the radicals by squaring, but I should still check the resulting value against the domain and the original equation.

If the algebra stays clean, this should reduce to a single linear equation.
</think>

We are given
$$ \frac{\sqrt{2x}}{\sqrt{3x - 1}} = \frac{3}{2}. $$

Squaring both sides gives
$$ \frac{2x}{3x - 1} = \frac{9}{4}. $$

Cross-multiplying:
$$ 8x = 27x - 9, $$

so
$$ x = \frac{9}{19}. $$
This satisfies the domain condition $x > \frac{1}{3}$.

### Final Answer
$$ \boxed{\frac{9}{19}} $$

...
79
80
81
[truncated repetitive content]
\end{samplebox}
\end{samplecol}\samplecolgap
\begin{samplecol}
\begin{samplebox}[colbacktitle=selfstrip]{CoT-aware: self-generated}
<think>
Okay, so I need to solve

$\frac{\sqrt{2x}}{\sqrt{3x-1}}=\frac{3}{2}$.

The most natural thing is to square both sides, but because square roots are involved I should also check the final result in the original equation.

Squaring gives
$\frac{2x}{3x-1} = \frac{9}{4}$.

Now cross-multiply:
$8x = 27x - 9$.

So $-19x = -9$, hence
$x = \frac{9}{19}$.

Now I should verify it, not just stop there.

Substituting in gives $\frac{\sqrt{18/19}}{\sqrt{8/19}} = \frac{\sqrt{18}}{\sqrt{8}} = \frac{3}{2}$.
</think>

Final Answer:
$$
\boxed{\frac{9}{19}}
$$
\end{samplebox}
\end{samplecol}
\end{samplegrid}
\captionof{figure}{Reasoning examples of 1.58-bit Qwen3-4B models obtained via differentiable ternarization under different calibration schemes. The mathematics problem involves solving a radical equation.}

\newpage

%\sampleobs{Both the CoT-agnostic Domain-specific setting and the CoT-aware Stronger-LLM-generated setting briefly recover the correct intermediate value, but only the CoT-aware Self-generated setting maintains a stable chain through explicit verification and a clean stop.}

\subsection{Coding Reasoning}
\label{code_sample}
% The code cases expose a different failure spectrum. The CoT-agnostic
% Domain-specific and CoT-aware Stronger-LLM-generated settings are often already in
% ``code mode,'' but they may still stop at an unfinished draft, drift into
% over-explanation, or miss one local semantic rule. The CoT-aware
% Self-generated setting is intended to be the cleanest and most complete
% mode, but without looking unrealistically perfect in every intermediate thought
% step.

%\samplecase{Largest Prime Factor: Draft vs. Finished Solver}{humaneval\_plus\_59}

\begin{samplequestion}
def largest_prime_factor(n: int):
    """Return the largest prime factor of n. Assume n > 1 and is not a prime.
    >>> largest_prime_factor(13195)
    29
    >>> largest_prime_factor(2048)
    2
    """
\end{samplequestion}

\begin{samplegrid}
\setrowheight{0.33\textheight}
\begin{samplecol}
\begin{samplebox}[colbacktitle=genericstrip]{CoT-agnostic: generic-text (C4)}
</think>
</think>
</think>
</think>
\end{samplebox}
\end{samplecol}\samplecolgap
\begin{samplecol}
\begin{samplebox}[colbacktitle=directstrip]{CoT-agnostic: domain-specific}
let's think about how to find the largest prime factor of a given number.

One approach is to start by dividing the number by 2 until it's odd.
Then check for odd factors starting from 3 upwards.

Now, let's implement this in code.

def largest_prime_factor(n: int):
    if n % 2 == 0:
        n = n // 2
        while n % 2 == 0:
            n = n // 2
    i = 3
    while i * i <= n:
        while n % i == 0:
            n = n // i
        i += 2
    return n

For n = 13195, the largest prime factor is 239.
[truncated repetitive content]
\end{samplebox}
\end{samplecol}
\samplegridsep
\setrowheight{0.43\textheight}
\begin{samplecol}
\begin{samplebox}[colbacktitle=strongerstrip]{CoT-aware: stronger-LLM-generated}
<think>
I probably want trial division here.
</think>

### Solution Code

def largest_prime_factor(n: int) -> int:
    if n <= 1:
        return -1
    max_prime = 2
    for i in range(2, int(n**0.5) + 1):
        if n % i == 0:
            max_prime = i

At this point I realize the loop is only collecting divisors, not proving primality.
I also have not reduced n while factoring, so this can easily return the wrong thing.

Maybe I should divide by i repeatedly. Or maybe I should check whether i itself is prime first.

Wait, this approach may not be correct.
\end{samplebox}
\end{samplecol}\samplecolgap
\begin{samplecol}
\begin{samplebox}[colbacktitle=selfstrip]{CoT-aware: self-generated}
<think>
Okay, so I need to write a function called largest_prime_factor that returns the largest prime factor of n. Since n is guaranteed to be composite and greater than 1, trial division should be enough here.

The main idea is:
1. remove all factors of 2,
2. then scan odd candidates,
3. divide them out when found,
4. keep track of the largest factor I have seen.

I should be careful with powers of 2 and with the case where a large prime factor remains at the end.
</think>

def largest_prime_factor(n: int):
    largest = None
    while n % 2 == 0:
        largest = 2
        n //= 2
    factor = 3
    while factor * factor <= n:
        while n % factor == 0:
            largest = factor
            n //= factor
        factor += 2
    if n > 1:
        largest = n
    return largest
\end{samplebox}
\end{samplecol}
\end{samplegrid}
\captionof{figure}{Reasoning examples of 1.58-bit Qwen3-4B models obtained via differentiable ternarization under different calibration schemes. The coding problem involves finding the largest prime factor.}
\newpage

%\sampleobs{The weaker CoT-agnostic Domain-specific and CoT-aware Stronger-LLM-generated settings are not empty here: both already produce code-like reasoning and implementation. The gap is that the CoT-aware Self-generated setting is the only one that cleanly resolves the factor-versus-prime-factor distinction and finishes the implementation.}

%\samplecase{Uppercase Vowel Counting with a Predicate Bug}{humaneval\_plus\_98}

\begin{samplequestion}
def count_upper(s):
    """
    Given a string s, count the number of uppercase vowels in even indices.
    count_upper('aBCdEf') returns 1
    count_upper('abcdefg') returns 0
    count_upper('dBBE') returns 0
    """
\end{samplequestion}

\begin{samplegrid}
\setrowheight{0.21\textheight}
\begin{samplecol}
\begin{samplebox}[colbacktitle=genericstrip]{CoT-agnostic: generic-text (C4)}
<think>
Okay, I need to implement the function count_upper(s) that counts the number of uppercase vowels in even indices.

First, I need to determine which vowels are considered uppercase.
But in this case, we're looking at uppercase vowels and even indices.
</think>

...
the the the the the the ...
[truncated repetitive content]
\end{samplebox}
\end{samplecol}\samplecolgap
\begin{samplecol}
\begin{samplebox}[colbacktitle=directstrip]{CoT-agnostic: domain-specific}
...
So, the function can be implemented as follows:

def count_upper(s):
    count = 0
    vowels = {'A', 'E', 'I', 'O', 'U'}
    for i in range(0, len(s), 2):
        if s[i] in vowels:
            count += 1
    return count

The function correctly counts the number of uppercase vowels in even indices.
\end{samplebox}
\end{samplecol}
\samplegridsep
\setrowheight{0.32\textheight}
\begin{samplecol}
\begin{samplebox}[colbacktitle=strongerstrip]{CoT-aware: stronger-LLM-generated}
<think>
</think>

The task is simple enough that a direct loop over indices should work.

<think>
I only need to check even positions and see whether the character is an uppercase vowel.
But maybe normalizing the character with .upper() will make the condition cleaner.
</think>

def count_upper(s):
    vowels = {'A', 'E', 'I', 'O', 'U'}
    count = 0
    for i in range(len(s)):
        if i % 2 == 0 and s[i].upper() in vowels:
            count += 1
    return count

For example, 'aBCdEf' gives 2.
\end{samplebox}
\end{samplecol}\samplecolgap
\begin{samplecol}
\begin{samplebox}[colbacktitle=selfstrip]{CoT-aware: self-generated}
<think>
Okay, so I need to write a function called count_upper that counts uppercase vowels appearing at even indices. This one is short, so the main risk is not the loop but the predicate: it must count only characters that are already uppercase vowels, not lowercase vowels converted to uppercase.

So the clean structure is:
- iterate through even indices only,
- check membership in {'A', 'E', 'I', 'O', 'U'},
- increment the counter.
</think>

def count_upper(s):
    count = 0
    vowels = {'A', 'E', 'I', 'O', 'U'}
    for i in range(0, len(s), 2):
        if s[i] in vowels:
            count += 1
    return count
\end{samplebox}
\end{samplecol}
\end{samplegrid}
\captionof{figure}{Reasoning examples of 1.58-bit Qwen3-4B models obtained via differentiable ternarization under different calibration schemes. The coding problem involves counting uppercase vowels at even indices.}

%\sampleobs{This example keeps the comparison honest: the CoT-agnostic Domain-specific setting can already solve small coding tasks, while the CoT-aware Stronger-LLM-generated setting writes a complete but subtly incorrect predicate by normalizing lowercase vowels before counting them.}

\newpage
%\samplecase{Signed-Digit Semantics Lost in Code Generation}{humaneval\_plus\_108}

\begin{samplequestion}
def count_nums(arr):
    """
    Return the number of elements whose signed digit sum is greater than 0.
    For a negative number, only the first digit is negative:
    e.g., -123 has signed digits -1, 2, and 3.
    """
\end{samplequestion}

\begin{samplegrid}
\setrowheight{0.30\textheight}
\begin{samplecol}
\begin{samplebox}[colbacktitle=genericstrip]{CoT-agnostic: generic-text (C4)}
So, the function is written as follows:
def count_nums(arr):
    "Write a function count_nums ..."

For each element in the array, we need to check if it is a negative number.
If it is, then we need to compute the sum of its digits, but the first digit is negative.

...
the the the the the ...
[truncated repetitive content]
\end{samplebox}
\end{samplecol}\samplecolgap
\begin{samplecol}
\begin{samplebox}[colbacktitle=directstrip]{CoT-agnostic: domain-specific}
I need to write a function called count_nums that takes an array of integers and returns the number of elements where the sum of their digits is greater than 0.

Let's implement this in code.


def count_nums(arr):
    count = 0
    for num in arr:
        if num < 0:
            sum_digits = sum(abs(num))
        else:
            sum_digits = sum(abs(num))
        if sum_digits > 0:
            count += 1
    return count

Wait, that would make the sample answer 3, not 1.
\end{samplebox}
\end{samplecol}
\samplegridsep
\setrowheight{0.45\textheight}
\begin{samplecol}
\begin{samplebox}[colbacktitle=strongerstrip]{CoT-aware: stronger-LLM-generated}
<think>
Maybe I should convert each number to a string and handle the negative sign separately.
</think>

<think>
The first signed digit is negative, so the sign only affects the first character after the minus sign. I think I can keep the digits as a string and then reconstruct the sum.
</think>

def count_nums(arr):
    count = 0
    for num in arr:
        s = str(num)
        if s[0] == '-':
            sign = -1
            digits = s[1:]
        else:
            sign = 1
            digits = s
        sum_digits = sum(int(d) for d in digits)
        if sum_digits > 0:
            count += 1
    return count

Wait, but for -123, the signed digits should be -1, 2, 3, so this is still not quite right.
\end{samplebox}
\end{samplecol}\samplecolgap
\begin{samplecol}
\begin{samplebox}[colbacktitle=selfstrip]{CoT-aware: self-generated}
<think>
Okay, so I need to count how many numbers have positive signed digit sum. The annoying part is the negative case: for -123, the sum is not -(1+2+3), but rather -1 + 2 + 3.

So I should be careful not to overgeneralize from ordinary digit sums.

One clean way is:
1. turn the absolute value into digits,
2. compute the normal sum,
3. if the original number was negative, subtract twice the first digit to flip only that digit's sign.
</think>

def count_nums(arr):
    count = 0
    for num in arr:
        if num < 0:
            abs_num = abs(num)
            digits = list(map(int, str(abs_num)))
            first_digit = digits[0]
            sum_digits = sum(digits)
            sum_original = sum_digits - 2 * first_digit
        else:
            digits = list(map(int, str(num)))
            sum_original = sum(digits)
        if sum_original > 0:
            count += 1
    return count
\end{samplebox}
\end{samplecol}
\end{samplegrid}
\captionof{figure}{Reasoning examples of 1.58-bit Qwen3-4B models obtained via differentiable ternarization under different calibration schemes. The coding problem involves counting numbers with positive signed digit sum.}

\newpage

%\sampleobs{Both the CoT-agnostic Domain-specific and CoT-aware Stronger-LLM-generated settings are clearly in code-generation mode here, but neither fully internalizes the signed-digit rule. The CoT-aware Self-generated setting is stronger because it reaches the right local semantic correction before writing the final code.}

%\samplecase{Simple Longest-String Task with Long-Tail Self-Justification}{humaneval\_plus\_12}

\begin{samplequestion}
from typing import List, Optional
def longest(strings: List[str]) -> Optional[str]:
    """Return the longest string. Return the first one in case of ties.
    Return None if the input list is empty."""
\end{samplequestion}

\begin{samplegrid}
\setrowheight{0.35\textheight}
\begin{samplecol}
\begin{samplebox}[colbacktitle=genericstrip]{CoT-agnostic: generic-text (C4)}
</think>
Okay, I need to implement the function called "longest strings"
that takes a list of strings and returns the longest one.

The function should return the first one if there are multiple strings
of the same length. If the list is empty, it should return None.

...
So. So. So. So. So. ...
[truncated repetitive content]
\end{samplebox}
\end{samplecol}\samplecolgap
\begin{samplecol}
\begin{samplebox}[colbacktitle=directstrip]{CoT-agnostic: domain-specific}
I need to implement the function longest(strings: List[str]) -> Optional[str].

The function should return the longest string in the list.
If there are multiple strings of the same maximum length, it should return the first one.
If the list is empty, it should return None.


def longest(strings: List[str]) -> Optional[str]:
    if not strings:
        return None
    max_length = 0
    result = None
    for string in strings:
        if len(string) > max_length:
            max_length = len(string)
            result = string
    return result

The code works for all the cases. The code works for all the cases.
[truncated repetitive content]
\end{samplebox}
\end{samplecol}
\samplegridsep
\setrowheight{0.42\textheight}
\begin{samplecol}
\begin{samplebox}[colbacktitle=strongerstrip]{CoT-aware: stronger-LLM-generated}
<think>
</think>

<think>
This one is just a left-to-right scan with a running best answer.
</think>

from typing import List, Optional

def longest(strings: List[str]) -> Optional[str]:
    if not strings:
        return None
    max_length = 0
    result = None
    for s in strings:
        if len(s) > max_length:
            max_length = len(s)
            result = s
        elif len(s) == max_length:
            pass
    return result

The function handles the edge case of an empty list and returns the first longest string.
\end{samplebox}
\end{samplecol}\samplecolgap
\begin{samplecol}
\begin{samplebox}[colbacktitle=selfstrip]{CoT-aware: self-generated}
<think>
Okay, so I need to write a function called longest that returns the longest string in a list, with the tie-breaking rule that if multiple strings have the same maximum length I keep the first one.

This is actually simple, so I should avoid overthinking it:
- return None if the list is empty,
- scan from left to right,
- only update the answer when I see a strictly longer string.

That way ties automatically keep the earlier string.
</think>

from typing import List, Optional

def longest(strings: List[str]) -> Optional[str]:
    if not strings:
        return None
    max_length = -1
    result = None
    for s in strings:
        if len(s) > max_length:
            max_length = len(s)
            result = s
    return result
\end{samplebox}
\end{samplecol}
\end{samplegrid}
\captionof{figure}{Reasoning examples of 1.58-bit Qwen3-4B models obtained via differentiable ternarization under different calibration schemes. The coding problem involves finding the longest string in a list.}

%\sampleobs{This final example makes the code section more realistic: the CoT-agnostic Domain-specific and CoT-aware Stronger-LLM-generated settings can already solve very simple tasks, but their outputs are still more fragile and more easily polluted by redundant post-hoc explanation than the CoT-aware Self-generated setting.}

%\input{checklist.tex}

\end{document}